\documentclass[lettersize,journal]{IEEEtran}
\usepackage{amsmath,amsfonts}
\usepackage{array}
\usepackage{subcaption}
\usepackage{textcomp}
\usepackage{stfloats}
\usepackage{url}
\usepackage{verbatim}
\usepackage{graphicx}
\usepackage{cite}
\usepackage{amsmath,amssymb,amsfonts}
\usepackage{bbm}
\usepackage{makecell}
\DeclareMathOperator*{\argmax}{arg\,max}

\usepackage{pifont}
\usepackage{algorithm}
\usepackage{algorithmic}

\usepackage{graphicx}
\usepackage{textcomp}
\usepackage{stfloats}
\usepackage{mathtools, cuted}
\usepackage{amsthm}
\newtheorem{theorem}{Theorem}
\newtheorem{lemma}{Lemma}
\newtheorem{assumption}{Assumption}
\DeclareMathOperator\erf{erf}
\usepackage{xcolor}
\usepackage{array}
\usepackage{booktabs}
\usepackage{mathtools}
\usepackage{tabularray}
\usepackage{tabularx}
\usepackage{enumitem}
\usepackage{caption}
\usepackage{csquotes}
\usepackage[font=small]{caption}
\usepackage[normalem]{ulem}
\usepackage{cancel}

\begin{document}
\title{Robust Bayesian Optimization via \\ Localized Online Conformal Prediction\\

\author{
{Dongwon Kim,~\IEEEmembership{Student Member,~IEEE,}
Matteo Zecchin,~\IEEEmembership{Member,~IEEE},\\
Sangwoo Park,~\IEEEmembership{Member,~IEEE},
Joonhyuk Kang,~\IEEEmembership{Member,~IEEE}, and \\
Osvaldo Simeone,~\IEEEmembership{Fellow,~IEEE}}
}

\thanks{D. Kim and J. Kang are with the School of Electrical Engineering, Korea Advanced Institute of Science and Technology (KAIST), Daejeon 34141, South Korea (e-mail: davinci5694@kaist.ac.kr; jhkang@ee.kaist.ac.kr).}
\thanks{M. Zecchin, and O. Simeone are with the King's Communications, Learning \& Information Processing (KCLIP) lab within the Centre for Intelligent Information Processing Systems (CIIPS), Department of Engineering, King's College London, W2CR 2LS London, U.K. (e-mail: \{matteo.1.zecchin, osvaldo.simeone\}@kcl.ac.uk).}
\thanks{S. Park is with the Department
of Electrical and Electronic Engineering, Imperial College London, London SW7 2AZ, UK. (e-mail: s.park@imperial.ac.uk).}
\thanks{This research was supported by the MSIT (Ministry of Science and ICT), Korea, under the ITRC (Information Technology Research Center) support program (IITP-2025-2020-0-01787) supervised by the IITP (Institute of Information \& Communications Technology Planning \& Evaluation)}
\thanks{The work of M. Zecchin and O. Simeone is supported by the European Union’s Horizon Europe project CENTRIC (101096379). The work of O. Simeone is also supported by an Open Fellowship of the EPSRC (EP/W024101/1), and by the EPSRC project (EP/X011852/1).}
\thanks{The authors would like to thank Dr. Nicola Paoletti (King's College London) for early discussions that motivated parts of this research.}
}
\maketitle

\begin{abstract}
Bayesian optimization (BO) is a sequential approach for optimizing black-box objective functions using zeroth-order noisy observations. In BO, Gaussian processes (GPs) are employed as probabilistic surrogate models to estimate the objective function based on past observations, guiding the selection of future queries to maximize utility. However, the performance of BO heavily relies on the quality of these probabilistic estimates, which can deteriorate significantly under model misspecification. To address this issue, we introduce localized online conformal prediction-based Bayesian optimization (LOCBO), a BO algorithm that calibrates the GP model through localized online conformal prediction (CP). LOCBO corrects the GP likelihood based on predictive sets produced by LOCBO, and the corrected GP likelihood  is then denoised to obtain a calibrated posterior distribution on the objective function. The likelihood calibration step leverages an input-dependent calibration threshold to tailor coverage guarantees to different regions of the input space. Under minimal noise assumptions, we provide theoretical performance guarantees for LOCBO’s iterates that hold for the unobserved objective function. These theoretical findings are validated through experiments on synthetic and real-world optimization tasks, demonstrating that LOCBO consistently outperforms state-of-the-art BO algorithms in the presence of model misspecification.

\end{abstract}


\section{Introduction}\label{sec:intro}

\subsection{Motivation}
\label{sec:intro_motivation}
{Black-box optimization} refers to a class of optimization problems in which information about the objective function can be only obtained through sequential queries to a zeroth-order oracle. In these problems, the goal is to minimize the number of queries necessary to find a good approximation of the objective function’s maximizer. \textit{Bayesian optimization} (BO) is a popular tool for black-box optimization that provides a principled approach to account for uncertainty about the objective function \cite{wang2022tight, zhang2023bayesian}. In BO, a Gaussian probabilistic model is used to estimate the objective function based on the available, noisy, function evaluations, guiding decisions on which candidate solutions to query next.

BO assumes a \textit{Gaussian process} (GP) prior on the space of objective function and a Gaussian likelihood. Under these assumptions, a tractable posterior distribution for the objective function is derived by using Bayes rule. Therefore, the performance of BO hinges on the validity of the assumed model for the objective function and for the observation noise. However, in real-world problems, the assumed
model is often \textit{misspecified}. In particular, the true objective function may not match the smoothness level assumed by the GP \cite{snoek2014input, bodin2020modulating, bogunovic2021misspecified}, and the observation noise may be non-Gaussian or have
input-dependent variance \cite{griffiths2021achieving, 9212578}. These discrepancies between model and ground-truth objective and observation noise can ultimately hinder the performance of BO \cite{schulz2016quantifying, shahriari2015taking}. The reader is referred to \cite[Figure 1]{neiswanger2021uncertainty} for an example.  

This paper aims at developing GP-based BO strategies that offer performance guarantees irrespective of prior misspecification and under minimal assumptions on the observation noise.
\vspace{-11pt}
\begin{figure*}[t]
\centering
\centerline{\includegraphics[width=0.72\textwidth]{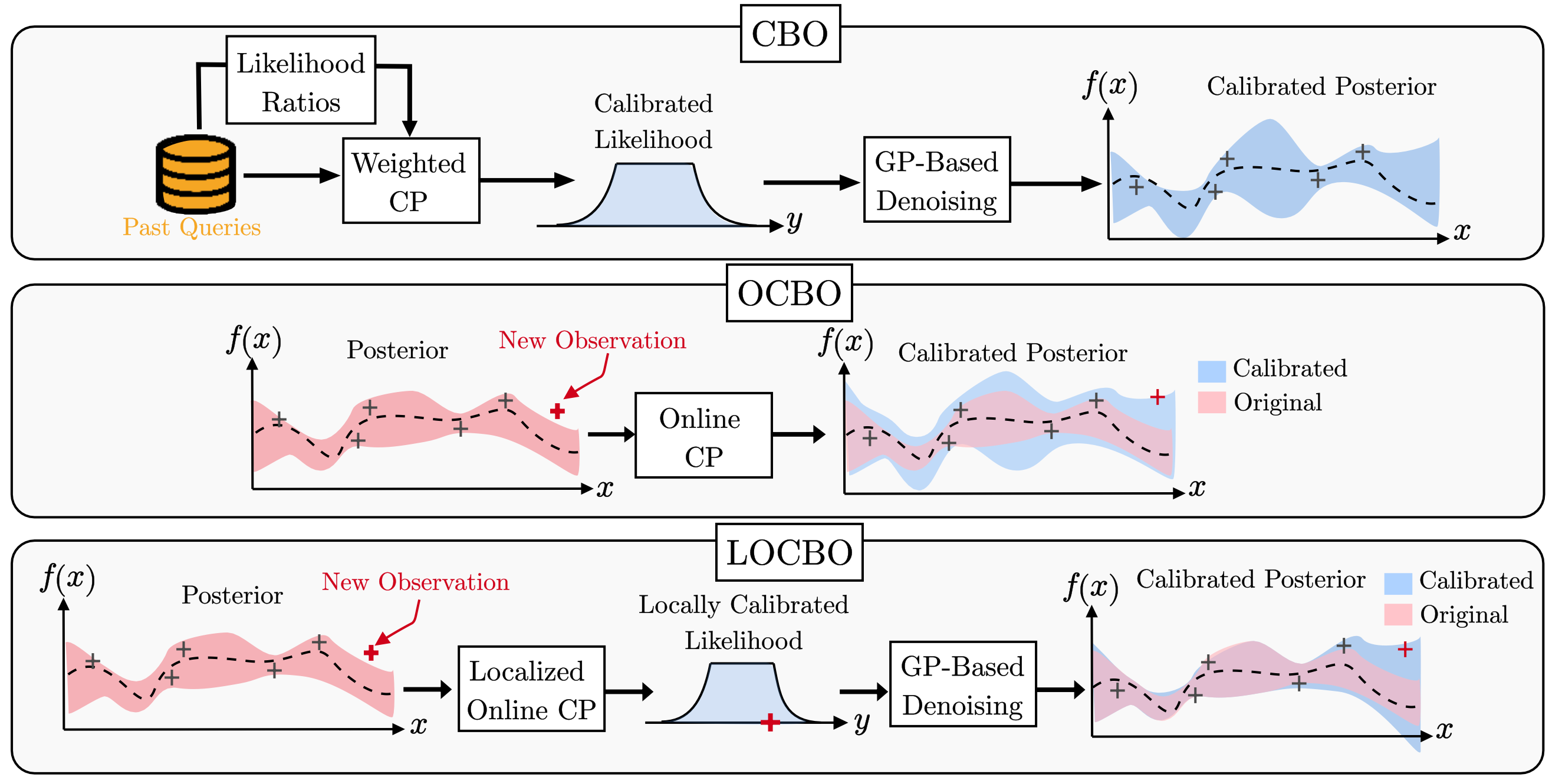}}
\caption{ (Top) Offline CP-based Bayesian Optimization (CBO) \cite{stanton2023bayesian} leverages past queries to calibrate the likelihood for the noisy observations $y$, which is then ``denoised'' to obtain a calibrated posterior for the objective function $f(\mathbf{x})$. (Middle) OCBO \cite{deshpande2024online} uses online CP to directly calibrate the posterior distribution of the objective function $f(\mathbf{x})$ based on the feedback of past queries, while providing performance guarantees in the presence of noiseless observations. (Bottom) The proposed localized online CP-based Bayesian Optimization (LOCBO) uses localized online CP \cite{zecchin2024localized} to calibrate locally the likelihood function, which is then ``denoised'' to compute a calibrated posterior distribution over the objective function $f(\mathbf{x})$. LOCBO provides performance guarantees irrespective of prior misspecification and under minimal assumptions on the observation noise.}
\label{fig:comp_methods}
\vspace{-12pt}
\end{figure*}
\subsection{State of the Art}

To mitigate the outlined problem, several solutions have been investigated. Non-stationary kernels and multi-layer GPs have been used to enhance the expressive power of conventional GPs \cite{albahar2022physics, marmin2018warped}. Prior misspecification can be further addressed via meta-learning \cite{nikoloska2022bayesian, volpp2019meta, rothfuss2023meta} if one has access to data from optimization problem instances that share similar objective functions. Meta-learning methods optimize the kernel function to tailor the GP prior to the given class of optimization problems. Also related is work on active learning with prior or model misspecifications \cite{sugiyama2005active, fudenberg2017active}. However, active learning and BO have very different goals: BO addresses black-box optimization, while active learning tackles the optimization of predictive models focusing on generalization \cite{svendsen2020active, fernandez2020adaptive}. 

All the approaches reviewed above lack formal performance guarantees in the presence of model misspecification. In contrast, confidence sequences were used in \cite{neiswanger2021uncertainty} to obtain statistically valid confidence intervals on the objective function value. However, this work makes the simplifying assumption that the objective function belongs to a pre-specified set of possible functions.

More recently, \textit{conformal prediction} (CP) \cite{vovk2005algorithmic} was introduced as an assumption-free method to recalibrate the uncertainty estimates provided by GP regression in BO. In the context of CP, calibration refers to the property that the predicted confidence sets (or intervals) contain the true outcomes with the specified probability \cite{angelopoulos2021gentle}. CP is a \textit{post-hoc calibration} method that can be used to extract prediction sets with coverage guarantees from the decisions of an arbitrary pre-trained model. CP can operate offline or online. \textit{Offline CP} works on a fixed calibration dataset that is assumed to follow the same distribution of the test data \cite{vovk2005algorithmic, angelopoulos2021gentle}. In contrast, \textit{online CP} is formulated in an online sequential setting, in which decisions are made over time and feedback is available after each decision \cite{gibbs2021adaptive, angelopoulos2024online, feldman2022achieving}.

Offline CP has been applied to BO to calibrate the GP's posterior distribution on the objective function \cite{stanton2023bayesian}. As illustrated in the top panel of Figure \ref{fig:comp_methods}, \textit{offline CP-based BO} (CBO) leverages offline CP \cite{vovk2005algorithmic} to calibrate the likelihood of the noisy observations $y$ based on previous noisy evaluations of the objective function. In particular, CBO adopts weighted CP  (WCP) \cite{tibshirani2019conformal}, a variant of offline CP,  to account for the distribution shift between past and future input-output pairs that is caused by the adaptive data acquisition strategy implemented by BO. The calibrated likelihood is then ``denoised'' by using the GP model to obtain a calibrated posterior on the objective function $f(\mathbf{x})$.  

Importantly, given its reliance on WCP, CBO requires access to likelihood ratios, accounting for distributional shifts in the optimization process, which are generally difficult to obtain and can only be approximated. To circumvent this problem, online CP has been used to directly calibrate the posterior distribution over the objective function $f(\mathbf{x})$, without first calibrating the likelihood of the noisy observations \cite{deshpande2024online}. As shown in the middle panel of Figure \ref{fig:comp_methods}, after observing the result of a query, OCBO adapts the posterior over the objective function $f(\mathbf{x})$.  OCBO performance guarantees hold in the absence of observation noise, leaving open the problem of designing methods that offer provable reliability measures in the presence of noisy observations and a mismatched likelihood.

While using different CP schemes, both CBO and OCBO target \textit{marginal} coverage guarantees. This means that, in the case of i.i.d data, their calibration guarantees hold \textit{on average} with respect to the distribution over the candidate inputs. Approaches designed for marginal calibration can lead to uncertainty quantification with uneven performance across the input space, where some parts of the input space are characterized by overly conservative uncertainty estimates, while other parts of the input space are assigned consistently invalid predicted sets \cite{guan2019conformal,gibbs2022conformal,hore2023conformal, zecchin2024localized}. This may negatively impact the performance of BO, which aims at identifying specific input areas that are likely to include optimal solutions.

This problem is addressed by partitioned GP and BO approaches \cite{lee2017hierarchically, eriksson2019scalable}, which split the input space into predefined regions, modeling the objective function in each region separately. However, like BO, partitioned BO does not offer any theoretical guarantees in the presence of model misspecification.

\subsection{Main Contributions}

In light of the limitations of existing approaches, we introduce \textit{localized online CP-based BO} (LOCBO). As shown in Figure \ref{fig:sota_comparison}, LOCBO borrows from CBO the idea of calibrating the likelihood function for the noisy observation $y$, which is converted into a posterior distribution over the objective function $f(\mathbf{x})$ via a denoising step. However, unlike CBO, LOCBO applies online CP for the calibration of the likelihood function, guaranteeing coverage properties without the need to make unrealistic assumptions about the availability of precise likelihood ratios. Furthermore, unlike OCBO, by incorporating an efficient denoising step, LOCBO offers theoretical guarantees also in the presence of noisy observations. 

As another innovation, LOCBO integrates a recently introduced localized version of online CP \cite{zecchin2024localized}. This approach allows LOCBO to tailor the calibration process to the properties of the objective function in the input space,  ensuring a more homogeneous level of coverage as a function of the optimization variables.

The main contributions of this work are summarized as follows:

\noindent $\bullet$ We study the problem of BO under model misspecification and propose LOCBO, an online calibration scheme based on localized online CP \cite{zecchin2024localized}.  As illustrated in Figure \ref{fig:comp_methods}, LOCBO corrects GP posterior estimates based on noisy queries by first calibrating the GP likelihood using localized online CP and then applying a denoising step. This way, LOCBO retains the simplicity of online calibration \cite{deshpande2024online}, while targeting localized calibration and accounting for noisy observation via denoising \cite{stanton2023bayesian}.

\noindent $\bullet$ We analyze the theoretical properties of LOCBO and demonstrate that, under minimal assumptions about the observation noise (Assumption \ref{ass:balanced_noise}), LOCBO yields posterior estimates that achieve long-term calibration for the unknown objective function values (Lemma \ref{lemma:long-run_obj}). This guarantee is shown to translate into a performance certificate for the utility of the iterates generated by LOCBO (Theorem \ref{th:utility_guarantee}). These conclusions extend existing results on online CP-based calibration, which were previously limited to noiseless cases \cite{deshpande2024online}.

\noindent $\bullet$ We benchmark the performance of LOCBO on optimization problems involving both synthetic objective functions \cite{surjanovic2013virtual} and real-world objectives \cite{benzaghta2023designing}. Empirical results indicate that the proposed scheme outperforms existing CP-based BO calibration algorithms. Furthermore, by analyzing the performance of LOCBO across different localization levels, we highlight the important role played by localized calibration in BO.

\begin{figure}[t]
    \centering
    \large
    \resizebox{\columnwidth}{!}{ 
    \begin{tabular}{>{\centering\arraybackslash}m{2cm}!{\vrule width 1pt}>{\centering\arraybackslash}m{4.4cm}|>{\centering\arraybackslash}m{4.9cm}|}
        \textbf{Method} & \textbf{Calibration Algorithm} & \textbf{Coverage Guarantee} \\ \Xhline{1.5pt}
        \vspace{0.5cm} CBO \cite{stanton2023bayesian} & 
        \vspace{0.5cm}Offline Weighted CP & 
        \begin{tabular}[c]{@{}c@{}}approximate, \\ probabilistic, \\ marginal\end{tabular} \\ \hline
        \vspace{0.4cm} OCBO \cite{deshpande2024online} & 
        \vspace{0.4cm} Online CP & 
        \vspace{-0.2cm}\begin{tabular}[c]{@{}c@{}}exact (noiseless observation), \\ deterministic, \\ marginal (for i.i.d. data)\end{tabular} \\ \hline
        \vspace{0.4cm} LOCBO (Proposed) & 
       \vspace{0.4cm} Localized Online CP & 
        \vspace{-0.2cm}\begin{tabular}[c]{@{}c@{}}exact, \\ deterministic, \\ localized (for i.i.d. data)\end{tabular} \\ 
    \end{tabular}
    }
    \caption{Comparison between state-of-the-art calibration-based BO schemes CBO \cite{stanton2023bayesian}, OCBO \cite{deshpande2024online}, and the proposed method LOCBO.}
    \label{fig:sota_comparison}
    \vspace{-11pt}
\end{figure}
\subsection{Organization}
The remainder of this paper is organized as follows: Section \ref{sec:Prob_Def} formulates the black-box optimization problem and provides an overview of Gaussian Processes (GP), and Bayesian Optimization (BO). Section \ref{sec:Prior Art} reviews conformal prediction (CP) and existing calibration-based BO algorithms. In Section \ref{sec:LOCBO}, the proposed method, LOCBO, is introduced, and its theoretical guarantees are studied. Section \ref{sec:Exp_results} compares LOCBO with other approaches using synthetic objective functions and real-world applications. Finally, Section \ref{sec:Conclusion} summarizes the findings of this study.

\section{Problem Definition}
\label{sec:Prob_Def}
We consider a general black-box optimization problem
\begin{align}
\label{eq:Black box}
    \mathbf{x}^* \in \argmax_{\mathbf{x}\in \mathcal{X}} f(\mathbf{x}),
\end{align}
where $\mathcal{X}$ is a bounded subset of the $d$-dimensional vector space $\mathbbm{R}^d$, and ${f:}\, \mathcal{X} \rightarrow \mathbb{R}$ is an unknown black-box objective function. We study the standard setting in which we only have access to a noisy zeroth-order oracle for the black-box function, which, when queried with an input $\mathbf{x}$, returns the observation $y \in \mathcal{Y}$ given by
\begin{align}
\label{eq:noise_modelling}
   y = f(\mathbf{x}) + \xi(\mathbf{x}),
\end{align}
where $\xi(\mathbf{x})$ is input-dependent noise that is not necessarily Gaussian distributed.

To seek an approximate solution to the problem (\ref{eq:Black box}), typical black-box optimization algorithms adopt an iterative strategy. Accordingly, at each round $t$, the zeroth order oracle is queried with the input $\mathbf{x}_{t}$ producing the observation $y_t$. After $T$ iterations, based on the collected evidence 
\begin{align}
\label{eq: D_T}
\mathcal{D}_T = \{(\mathbf{x}_{t},y_{t})\}^T_{t=1},
\end{align}
the algorithm returns a candidate solution $\hat{\mathbf{x}}$ whose suboptimality gap is defined as
\begin{align}
\label{eq:suboptimality gap}
\Delta(\hat{\mathbf{x}})=f(\hat{\mathbf{x}})-f(\mathbf{x}^*).
\end{align}
The goal of black-box optimization is to minimize the number of queries needed to reach a target suboptimality gap.

\subsection{Gaussian Process}
\label{sec:GP}
A Gaussian process (GP) is a stochastic process that is used in BO as a prior in the space of objective functions $f(\mathbf{x})$. A GP models the joint distribution of any number of values of function $f(\mathbf{x})$ at different input values $\mathbf{x}$ as random variables following a multivariate Gaussian distribution \cite{williams2006gaussian}. Accordingly, a GP is specified by a mean function $\mu(\mathbf{x}): \mathcal{X}\rightarrow\mathbbm{R}$ and by a kernel function $k(\mathbf{x},\mathbf{x}'): \mathcal{X}\times \mathcal{X} \rightarrow \mathbbm{R}$, with the latter dictating the correlation between the values of function $f(\mathbf{x})$ at different input vectors $\mathbf{x}$. It is common for the mean function to be set to zero, i.e., $\mu(\mathbf{x}) = 0$, while typical choices for the kernel function include the square exponential kernel or Mat\'ern kernel \cite{williams2006gaussian}. 

When the likelihood $p(y|f(\mathbf{x}))$ is Gaussian, the  observation is given by
\begin{align}
    \label{eq: observations}
    y = f(\mathbf{x}) + z,
\end{align}
where  $z \sim \mathcal{N}(0, \sigma^2)$ represents additive Gaussian noise. Under the Gaussian likelihood \eqref{eq: observations}, given a data set of observations $\mathcal{D}_t = \{(\mathbf{x}_{\tau}, y_{\tau})\}_{\tau=1}^{t}$, the GP posterior distribution $p(f(\mathbf{x})|\mathcal{D}_t)$ at input $\mathbf{x}$ that has the following Gaussian distribution \cite{williams2006gaussian} 
\begin{align}\label{eq: GP} 
    p(f(\mathbf{x})|\mathcal{D}_t) = \mathcal{N}(\mu(\mathbf{x}|\mathcal{D}_t),\sigma^2(\mathbf{x}|\mathcal{D}_t)),
\end{align}
in which the mean and variance functions are
\begin{subequations} \label{eq:GP_post_mean_var}
\begin{align} 
\label{eq:GP_post_mean}
    \centering
    \mu(\mathbf{x}|\mathcal{D}_t) &= \mathbf{k}(\mathbf{x})^\top(\mathbf{K}(\mathbf{X})+\sigma^2\mathbf{I}_t)^{-1}\mathbf{y},\\ \label{eq:GP_post_var}
    \sigma^2(\mathbf{x}|\mathcal{D}_t) &= k(\mathbf{x},\mathbf{x}) - \mathbf{k}(\mathbf{x})^\top(\mathbf{K}(\mathbf{X})+\sigma^2\mathbf{I}_{t})^{-1}\mathbf{k}(\mathbf{x}),
\end{align}
\end{subequations}
where $\mathbf{y}=[y_1, \ldots, y_{t}]^\top$ is the $t\times1$ observation vector, $\mathbf{k}(\mathbf{x}) = [k(\mathbf{x}, \mathbf{x}_1),\ldots,k(\mathbf{x},\mathbf{x}_{t})]^\top$ is a $t\times 1$ covariance vector, and $\mathbf{K}(\mathbf{X})$ is a $t\times t$ covariance matrix, whose $(i,j)$-th entry is $ k(\mathbf{x}_i, \mathbf{x}_j)$. Furthermore, by the Gaussian likelihood \eqref{eq: observations} assumption, the distribution of $y$ follows Gaussian distribution
\begin{align}
    p(y|\mathbf{x}, \mathcal{D}_t) = \mathcal{N}(\mu(\mathbf{x}|\mathcal{D}_t),\sigma^2(\mathbf{x}|\mathcal{D}_t)+\sigma^2).
    \label{eq:GP_observation}
\end{align}

\subsection{Bayesian Optimization}
In conventional BO, a GP is used to model the distribution of the objective function $f(\mathbf{x})$, and a Gaussian likelihood $p(y|f(\mathbf{x}))$ as in \eqref{eq: observations} is used to model the observation $y$. At each optimization round $t$, based on the observations $\mathcal{D}_t$ in \eqref{eq: D_T}, the next candidate solution $\mathbf{x}_{t+1}$ is given by the maximizer of the acquisition function $a(\mathbf{x}|\mathcal{D}_t): \mathcal{X} \rightarrow \mathbbm{R}$,
\begin{align}
    \label{eq: argmax acquisition function}
    \mathbf{x}_{t+1} \in \argmax_{\mathbf{x}\in\mathcal{X}} a(\mathbf{x}| \mathcal{D}_t).
\end{align}
The acquisition function is derived from an \emph{utility function} $u(\mathbf{x}, f(\mathbf{x}), \mathcal{D}_t)$, which quantifies the benefit of evaluating the objective function $f(\mathbf{x})$ at the input $\mathbf{x}$, given the information provided by the dataset $\mathcal{D}_t$. Specifically, the acquisition function corresponds to the expected value of the utility function under the GP posterior distribution $p(f(\mathbf{x})|\mathcal{D}_t)$ \eqref{eq: GP},
\begin{align}
\label{eq:acqusition function}
a(\mathbf{x}|\mathcal{D}_t) = \int u(\mathbf{x}, f(\mathbf{x}), \mathcal{D}_t)p(f(\mathbf{x})|\mathcal{D}_t) \mathrm{d}f.
\end{align} 
The utility $u(\cdot,\cdot,\cdot)$ can be designed in different ways, aiming to strike distinct trade-offs between exploration and exploitation. Popular examples include \emph{expected improvement} (EI), the upper confidence bound (UCB) \cite{srinivas2009gaussian}, and entropy search (ES) \cite{wilson2017reparameterization}.
\section{Prior Art: Bayesian Optimization with Conformal Prediction}
\label{sec:Prior Art}
To mitigate the detrimental effects of model misspecification in BO, CP-based calibration was applied in prior work with the goal of improving the quality of the expected utility estimate \eqref{eq:acqusition function}. In this section, we review CBO \cite{stanton2023bayesian}, which uses WCP \cite{tibshirani2019conformal}, and OCBO \cite{deshpande2024online}, which employs online CP \cite{gibbs2021adaptive}.
 
\subsection{ Conformal Bayesian Optimization}
\label{sec:CBO}
In CBO \cite{stanton2023bayesian}, the GP posterior distribution $p(f(\mathbf{x})|\mathcal{D}_t)$ is replaced by
a posterior distribution $p_\alpha(f(\mathbf{x})|\mathcal{D}_t)$ that is calibrated via WCP \cite{tibshirani2019conformal}.

\subsubsection{Weighted Conformal Prediction}
WCP is a variant of CP that produces sets with valid coverage guarantees even in the presence of a covariate shift between calibration and test data. To describe it, fix two covariate distributions $p(\mathbf{x})$ and $p'(\mathbf{x})$ whose ratio 
\begin{align}
    w(\mathbf{x})=\frac{p'(\mathbf{x})}{p(\mathbf{x})}
\end{align}
is assumed to be known for all covariates $\mathbf{x}\in\mathbbm{R}^d$.

Given a \textit{calibration data set} $\mathcal{D}^{\text{cal}} = \{ (\mathbf{x}_i, y_i)\}_{i=1}^n$ with $(\mathbf{x}_i, y_i) \overset{\text{i.i.d.}}{\sim} p(\mathbf{x})p(y|\mathbf{x})$ for $i=1,...,n$, WCP produces prediction sets $\Gamma^{\rm WCP}_\alpha(\mathbf{x}) \subseteq \mathcal{Y}$ that include the test data point $(\mathbf{x}, y) \sim p'(\mathbf{x})p(y|\mathbf{x})$ with a user-defined probability level $1-\alpha$. Thus, while the conditional distribution $p(y|\mathbf{x})$ is assumed to be the same for both calibration and test data, the covariate distribution $p(\mathbf{x})$ and $p'(\mathbf{x})$ are generally distinct.

The prediction set $\Gamma^{\rm WCP}_\alpha(\mathbf{x}|\mathcal{D}^{\text{cal}})$ is obtained based on a \textit{non-conformity (NC) score} $s:\mathcal{X}\times\mathcal{Y}\to \mathbb{R}$, which quantifies the loss accrued by the model’s prediction for input $\mathbf{x}$ given the true label $y$. Furthermore, to account for the covariate shift, WCP weights the contribution of each calibration data point $(\mathbf{x}_i, y_i)$ using the ratio $w(\mathbf{x}_i)$. Accordingly, WCP evaluates the weighted empirical distribution over the NC scores given by
\begin{align}
   &p(s|\mathbf{x},\mathcal{D}^{\text{cal}})=  \frac{\sum^n_{i=1}w(\mathbf{x}_i)\delta_{s(\mathbf{x}_i,y_i)}(s)}{\sum^n_{i=1}w(\mathbf{x}_i)+w(\mathbf{x})}+w(\mathbf{x})\delta_{\infty}(s),
\end{align}
where $\delta_{x}(\cdot)$ is the Dirac delta function centered at $x$. For a target miscoverage level $\alpha\in[0,1]$, the corresponding WCP prediction set is finally given by
\begin{align}
    \Gamma^{\rm WCP}_\alpha(\mathbf{x}|\mathcal{D}^{\text{cal}}) = \left\{y \in \mathcal{Y} : s(\mathbf{x},y) \leq \mathcal{Q}_{1-\alpha} ( p(s|\mathbf{x},\mathcal{D}^{\text{cal}}) )  \right\},
    \label{eq:WCP_set}
\end{align}
where $\mathcal{Q}_{1-\alpha}(p(s))$ is the $(1-\alpha)$-quantile of the distribution $p(s)$.

WCP ensures the marginal coverage guarantee \cite{tibshirani2019conformal}
\begin{align}
    \Pr[y \in \Gamma^{\rm WCP}_\alpha(\mathbf{x}|\mathcal{D}^{\text{cal}})] \geq 1 - \alpha
    \label{eq:WCP_coverage}
\end{align}
where the probability is evaluated over the test input $(\mathbf{x},y)$ and over calibration data set $\mathcal{D}^{\text{cal}}$.
\subsubsection{Weighted Conformal Prediction-based Bayesian Optimization}

At each BO round $t$, CBO adopts WCP to calibrate the GP likelihood distribution $p(y|\mathbf{x},\mathcal{D}_t)$ based on the calibration data provided by the past queries $\mathcal{D}_t$. In this context, WCP is leveraged to account for the influence of the past queries on the next query, which causes a covariate shift (see, e.g., \cite[Section 2.1]{stanton2023bayesian}).  

\begin{figure}[t]
\centering
\centerline{\includegraphics[scale=0.95]{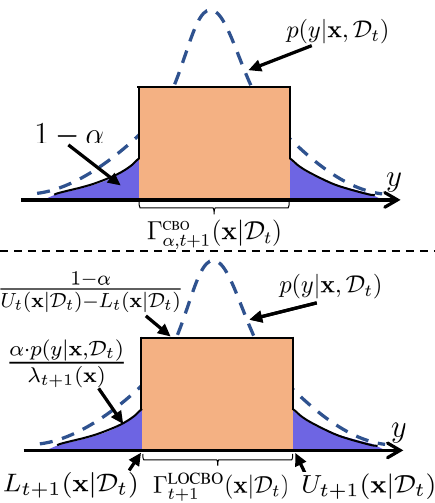}}
\caption{(top) Calibrated likelihood in CBO, which assumes a flat density within the prediction set $\Gamma_{\alpha, t+1}^{\text{\tiny{CBO}}}(\mathbf{x}|\mathcal{D}_t)$ returned by WCP, (bottom) Calibrated likelihood in LOCBO, which assumes a flat density within the prediction set $\Gamma^{\text{LOCBO}}_{t+1}(\mathbf{x}|\mathcal{D}_{t})$ obtained by localized online CP.}
\label{fig: CBO illustration}
\vspace{-0.6cm}
\end{figure}

Specifically, at each round $t$, CBO uses the available data $\mathcal{D}_t$, along with an estimate of the ratio $w(\mathbf{x})$, to produce the WCP set $ \Gamma^{\text{\tiny{CBO}}}_{\alpha, t+1}(\mathbf{x}|\mathcal{D}_t)$ using \eqref{eq:WCP_set}. As illustrated in the top part of Figure \ref{fig: CBO illustration}, CBO corrects the likelihood $p_\alpha^{\text{\tiny{CBO}}}(y|\mathbf{x},\mathcal{D}_t) $ so that the set $\Gamma_{\alpha, t+1}^{\text{\tiny{CBO}}}(\mathbf{x}|\mathcal{D}_{t})$ accounts for a probability $1-\alpha$ under the distribution $p_{\alpha}^{\text{\tiny{CBO}}}(y|\mathbf{x},\mathcal{D}_t)$. Based on the corrected likelihood $p_\alpha^{\text{\tiny{CBO}}}(y|\mathbf{x},\mathcal{D}_t) $, the calibrated posterior for the objective function is then obtained as
\begin{align}
	p_\alpha^{\text{\tiny{CBO}}}(f(\mathbf{x})|\mathcal{D}_t)\!\!=\!\!\!\!\int_{y'\in \mathcal{Y}}\!\!\!\!\!\! \!\!p(f(\mathbf{x})|\mathcal{D}_t\!\cup\!\{(\mathbf{x},y')\})p_\alpha(y'|\mathbf{x},\!\mathcal{D}_t)dy'\!,\!\!
 \label{eq:CBO_decomp}
\end{align}
where the marginalization over $y'$ effectively \enquote{denoises} the likelihood $p(y|\mathbf{x}, \mathcal{D}_t)$ by accounting for the GP estimate in \eqref{eq: GP}.
\vspace{-16pt}
\subsection{Online Conformal Bayesian Optimization}
\label{sec:OCBO}
The coverage guarantee of the prediction set $ \Gamma^{\text{\tiny{CBO}}}_{\alpha, t+1}(\mathbf{x}|\mathcal{D}_t)$ provided by CBO only holds under the unrealistic assumption that the weights $w(\mathbf{x})$ can be estimated precisely. To alleviate this problem,  OCBO \cite{deshpande2024online} adopts the online CP procedure \cite{gibbs2021adaptive} to directly calibrate the left quantile function associated with the GP posterior $p(f(\mathbf{x})|\mathcal{D}_t)$.  To elaborate, define the set of quantile levels $\mathcal{A}=\{\alpha_1,\dots,\alpha_M\}$ with $0\!\leq\!\alpha_i\!\leq\! 1$. Using the GP estimates in \eqref{eq:GP_observation}, OCBO produces at each time $t$ a quantile function $q_t(\alpha):[0,1]\to \mathcal{Y}$ that satisfies the long-run deterministic coverage property 
\begin{align} \label{eq:online_cal_marginal}
    \lim_{T \rightarrow \infty}\frac{1}{T}\sum_{t=1}^T \mathbbm{1}\{y_t \leq q_t\big(\alpha   \big)  \} = \alpha ,\text{ for all }\alpha\in \mathcal{A} 
\end{align}
 where $\mathbbm{1}\{\cdot\}$ equals 1 if the argument is true and equals 0 otherwise. This condition states that the long-term fraction of the time steps $t$ in which the observation $y_t$ does not exceed the quantile function $q_t(\alpha)$ tends to $\alpha$ for all levels $\alpha\in \mathcal{A}$. This ensures that the quantity  $q_t(\alpha)$ provides an asymptotically exact estimate of the $\alpha$-quantile of the observation $y_t$ in terms of the long-term average \eqref{eq:online_cal_marginal}.
{
\subsubsection{{Online Conformal Prediction}}
\label{sec:online_cp}
{Online CP is a calibration scheme that produces prediction sets with long-run deterministic coverage based on a data sequence $\{(\mathbf{x}_t,y_t)\}_{t\geq 1}$. Given a bounded NC score $s:\mathcal{X}\times\mathcal{Y}\to [0,B]$ and a fixed target miscoverage level $\alpha$, at every round $t$, online CP outputs the prediction set}
\begin{align}
    \label{eq:ocp_pred_set}{
    \Gamma^{\rm OCP}_{t}(\mathbf{x})=\left\{y:s(\mathbf{x},y)\leq \lambda_t \right\}.}
\end{align}
{The threshold $\lambda_t$, determining the size of the prediction set, is updated using the online update rule}
\begin{align}{
    \lambda_{t+1}=\lambda_{t}-\eta_t(\alpha-\mathbbm{1}\left\{y_{t}\notin\Gamma^{\rm OCP}_{t}(\mathbf{x}_{t}) \right\}),}
\end{align}
{with $\eta_t$ being a positive learning rate.  For a fixed learning rate $\eta_t=\eta$, the sequence of prediction sets $\Gamma^{\rm OCP}_{t}(\mathbf{x})$ satisfies the coverage guarantee~\mbox{\cite{gibbs2021adaptive}}}
\begin{align}
    \label{eq:locp_guarantee}
    {\frac{1}{T}\sum^T_{t=1}\mathbbm{1}\left\{y_t\in \Gamma^{\rm OCP}_{t}(\mathbf{x}_t)\right\}\geq 1-\alpha- \frac{B+\eta}{\eta T}}
\end{align}
{for any sequence $\{(\mathbf{x}_t,y_t)\}_{t\geq 1}$ and any time $T\geq 1$.}}

\subsubsection{Online Conformal Prediction-based Bayesian Optimization}

In OCBO, online CP is used to recalibrate the quantile function of the observation $y_t$ in terms of the GP likelihood $p(y|\mathbf{x},\mathcal{D}_t)$. Define as $Q_{\alpha}(p(y|\mathbf{x}, \mathcal{D}_t))$ the function returning the left $\alpha$-quantile of the GP likelihood $p(y|\mathbf{x}, \mathcal{D}_t)$ in \eqref{eq:GP_observation}. To obtain a sequence $q_t(\alpha)$ satisfying the calibration condition \eqref{eq:online_cal_marginal}, OCBO applies a recalibrator function $R_t: [0,1] \rightarrow [0,1]$ to the level $\alpha$ yielding a calibrated quantile function $q_t(\alpha) = \mathcal{Q}_{R_t(\alpha)}(p(y|\mathbf{x},\mathcal{D}_{t-1}))$. In practice, the recalibrator $R_t(\alpha)$ is optimized on a set of quantile levels $\mathcal{A}=\{\alpha_1,\dots,\alpha_M\}$, and approximated via linear interpolation for other values of probability $\alpha$. 

At each time $t$ and for each value $\alpha\in\mathcal{A}$, the prediction set is obtained from the calibrated quantile function as
\begin{align}
    \Gamma^{\text{OCBO}}_{t}(\mathbf{x}|\mathcal{D}_{t-1})\!=\!\left\{y:y \leq \mathcal{Q}_{R_{t}(\alpha)}(p(y|\mathbf{x},\mathcal{D}_{t-1}))\right\},
\end{align}
and the recalibrator is updated using the online CP rule
 \begin{align}
    \label{eq: OCBO update rule 1}
     R_{t+1}(\alpha)\! =\! R_{t}(\alpha)\! -\! \eta_{t}\big(\alpha\! -\! \mathbbm{1}\{y_t\notin \Gamma^{\text{OCBO}}_{t}(\mathbf{x}_t|\mathcal{D}_{t-1}) \}\big).
 \end{align}

Based on the guarantees of online CP, the recalibrated quantile can be shown to satisfy long-run deterministic coverage (\ref{eq:online_cal_marginal}) \cite[Theorem 2]{deshpande2024online}. In fact, for all $ \alpha \in\mathcal{A}$, the following limit holds
\begin{align}
    \label{eq: calibration condtion}
    \lim_{T\rightarrow \infty}\frac{1}{T}\sum_{t=1}^{T}\mathbbm{1}\{y_t\in \Gamma^{\text{OCBO}}_t(\mathbf{x}_t|\mathcal{D}_{t-1}) \} = \alpha.
    \end{align}

\section{Localized Online-Conformal-Prediction\\-based Bayesian Optimization}
\label{sec:LOCBO}

In this section, we introduce LOCBO, a novel CP-based calibration scheme for BO. Section \ref{sec:overview_locbo} provides an overview, and detailed descriptions are presented in Section \ref{sec:cal_likelihood_locbo}, \ref{sec:cal_likelihood_via_LOCP} and \ref{sec:posterior_calibration_locbo}. Theoretical calibration guarantees for LOCBO are proved in Section \ref{sec:theoretical_guarantees_locbo}.

\subsection{Overview}
\label{sec:overview_locbo}
LOCBO builds on two main ideas. As illustrated in Figure \ref{fig:comp_methods} and in the bottom part of Figure \ref{fig: CBO illustration}, the first idea is to leverage online CP instead of WCP as in CBO, to calibrate the likelihood $p(y|\mathbf{x}, \mathcal{D}_t)$. This makes it possible to benefit from assumption-free coverage guarantees for the predicted sets used to construct the likelihood $p(y|\mathbf{x}, \mathcal{D}_t)$. In contrast, as discussed in Section \ref{sec:CBO}, CBO's guarantees only hold under unrealistic assumptions about the quality of the estimation of the covariate likelihood ratios.

This approach makes it possible to operate in the presence of observation noise, unlike OCBO. In fact, the likelihood $p(y|\mathbf{x}, \mathcal{D}_t)$ can be \enquote{denoised} using minimal information about the observation noise \cite{stanton2023bayesian}, obtaining a calibrated posterior $p^{\text{\tiny{LOCBO}}}_\alpha(f(\mathbf{x})|\mathcal{D}_t)$ for the value of the objective function $f(\mathbf{x})$. 

The second main idea is to replace the standard online CP with localized online CP \cite{zecchin2024localized} in order to specialize the calibration of the likelihood $p(y|\mathbf{x}, \mathcal{D}_t)$ to different parts of the input space. As discussed in Section \ref{sec:intro}, this approach can tailor uncertainty estimates to the difficulty and data coverage of different regions of the input space.

The overall procedure is summarized in Algorithm \ref{alg:LOCBO}, and the next sections elaborate on the calibration of the likelihood and on denoising.
\vspace{-14pt}
\subsection{Localized Online CP}
\label{sec:cal_likelihood_locbo}
Localized online CP \cite{zecchin2024localized} is an online calibration scheme that generalizes the online CP, reviewed in Section \ref{sec:OCBO}, to calibrate set predictors so as to ensure long-run coverage guarantees. Unlike the online CP set in \eqref{eq:ocp_pred_set}, which is defined based on a scalar  $\lambda_t$, localized online CP produces prediction sets using the \textit{threshold function}
\begin{align}
    \label{eq:locbo_threshold}
    \lambda_t(\cdot)=g_t(\cdot)+c_t,
\end{align}
which comprises a constant $c_t$ and a function $g_t(\cdot)$. The function $g_t(\cdot)$ is constrained to belong to a reproducing kernel Hilbert space (RKHS) $\mathcal{H}$ specified by a kernel $k_g(\cdot,\cdot)$.

The kernel $k_g(\cdot,\cdot)$ determines the class of admissible threshold functions. This, in turn, dictates the degree of localization provided by localized CP \cite{zecchin2024localized}. For example, decreasing the length scale parameter $l$ in the radial basis function (RBF) kernel $k_g(\mathbf{x},\mathbf{x}')=\exp(-\lVert \mathbf{x}-\mathbf{x}'\rVert^2/l^2)$ increases the level of localization (see, \cite[Figure 2]{zecchin2024localized}).

The threshold function $\lambda_t(\cdot)$ in \eqref{eq:locbo_threshold} maps the input $\mathbf{x}$ to a localized threshold $\lambda_t(\mathbf{x})$ that is used to define the prediction set
\begin{align}
    \label{eq:locp_pred_set}
    \Gamma^{\rm LOCP}_{t}(\mathbf{x})=\left\{y:s_t(\mathbf{x},y)\geq \lambda_t(\mathbf{x})\right\}.
\end{align}
Note that, if we set $g_t(\cdot)=0$ for all $t\geq1$ in the threshold function \eqref{eq:locbo_threshold}, the prediction set \eqref{eq:locp_pred_set} coincides with the online CP prediction set \eqref{eq:ocp_pred_set}. Thus, the additional degree of choosing a threshold function $\lambda_t(\cdot)$ 
 in \eqref{eq:locp_pred_set} enables localized CP to tailor the set predictor to different regions of the input space.

In LOCBO, the localized CP set  \eqref{eq:locp_pred_set} is defined based on the conformity score
\begin{align}
    \label{eq:locbo_nc_score}
    s_t(\mathbf{x},y)=2 Q\left(\frac{|y-\mu(\mathbf{x}|\mathcal{D}_{t-1})|}{\tilde{\sigma}(\mathbf{x}|\mathcal{D}_{t-1})}\right),
\end{align}
where $Q(z)=\Pr[Z>z]$ is the complementary c.d.f. of the standard normal random variable $Z$, and $\mu(\mathbf{x}|\mathcal{D}_{t-1})$ and  $\tilde{\sigma}^2 (\mathbf{x}|\mathcal{D}_{t-1}) = {\sigma}^2 (\mathbf{x}|\mathcal{D}_{t-1})+ \sigma^2$ are the mean and variance of predictive posterior distribution \eqref{eq:GP_observation} at input $\mathbf{x}$.
Applying the non-conformity score \eqref{eq:locbo_nc_score} to the prediction set \eqref{eq:locp_pred_set} yields the interval
\begin{flalign}
\label{eq:locbo_prediction_set}
\Gamma_{t}^{\text{\tiny{LOCBO}}}(&\mathbf{x}|\mathcal{D}_{t-1})\hspace{-0.25em}=\hspace{-0.25em}\biggl\{y : \mu(\mathbf{x}|\mathcal{D}_{t-1})-Q^{-1}\hspace{-0.25em}\left(\hspace{-0.15em}\frac{\lambda_t(\mathbf{x})}{2}\hspace{-0.15em}\right)\tilde{\sigma}(\mathbf{x}|\mathcal{D}_{t-1})\nonumber\\& \leq \!y\!  \leq \mu(\mathbf{x}|\mathcal{D}_{t-1})+Q^{-1}\hspace{-0.25em}\left(\hspace{-0.15em}\frac{\lambda_t(\mathbf{x})}{2}\hspace{-0.15em}\right)\tilde{\sigma}(\mathbf{x}|\mathcal{D}_{t-1}) \biggr\}.
\end{flalign}
The set predictor $\Gamma_{t}^{\text{\tiny{LOCBO}}}(\mathbf{x}|\mathcal{D}_{t-1})$ returns intervals centered at the GP predictive posterior mean $\mu(\mathbf{x}|\mathcal{D}_{t-1})$ with a width that is proportional to the GP predictive posterior standard deviation $\tilde{\sigma}(\mathbf{x}|\mathcal{D}_{t-1})$. The input-dependent threshold $\lambda_t(\mathbf{x})$ determines the proportionality factor $Q^{-1}\left(\lambda_t(\mathbf{x})/2\right)$. We denote the upper and lower bounds of the LOCBO prediction set $\Gamma_{t}^{\text{\tiny{LOCBO}}}(\mathbf{x}|\mathcal{D}_{t-1})$ as $L_t(\mathbf{x}|\mathcal{D}_{t-1})$ and $U_t(\mathbf{x}|\mathcal{D}_{t-1})$, i.e., $\Gamma_{t}^{\text{\tiny{LOCBO}}}(\mathbf{x}|\mathcal{D}_{t-1})= [L_t(\mathbf{x}|\mathcal{D}_{t-1}), U_t(\mathbf{x}|\mathcal{D}_{t-1})]$ (see the bottom part of Figure \ref{fig: CBO illustration}).

At each round $t$, the observed query $(\mathbf{x}_t,y_t)$ is used to calibrate the threshold function $\lambda_{t}(\mathbf{x})$ in LOCBO's prediction set \eqref{eq:locbo_prediction_set}. The components of the function $\lambda_{t}(\mathbf{x})$ in \eqref{eq:locbo_threshold} are optimized online by using the update rule
\begin{subequations}
    \label{eq:locp_update_rule_const}
    \begin{align}
    c_{t+1}\! =&  c_{t} + \eta_{t}(\alpha - \mathbbm{1}\{y_t \notin  \Gamma_{t}^{\text{\tiny{LOCBO}}}(\mathbf{x}_t|\mathcal{D}_{t-1})\}), \\
   g_{t+1}(\cdot)\! =& (1-\lambda\eta_t)g_{t}(\cdot) + \eta_{t}(\alpha -  \mathbbm{1}\{y_t \notin  \Gamma_{t}^{\text{\tiny{LOCBO}}}(\mathbf{x}_t|\mathcal{D}_{t-1})\}) \nonumber\\
   &\cdot k_{g}(\mathbf{x}_t,\cdot),
   \end{align}
\end{subequations}
where $\eta_{t}>0$ is a learning rate, and $\lambda>0$ is a regularization hyperparameter using \eqref{eq:locp_update_rule_const}, constant $c_t$ is thus updated by following the online CP rule \eqref{eq:ocp_pred_set}, while the functional component $g_{t}(\cdot)$ is updated by using online kernel gradient descent \cite{kivinen2004online}.
\vspace{-5pt}
\subsection{Calibration of the Likelihood via Localized Online CP}
\label{sec:cal_likelihood_via_LOCP}
Based on the prediction set \eqref{eq:locbo_prediction_set}, LOCBO defines the calibrated likelihood
\begin{align} \label{eq:locbo_likelihood}
    p^{\text{\tiny{LOCBO}}}_\alpha(y|\mathbf{x},\mathcal{D}_t)= 
    \begin{cases}
        \frac{1-\alpha}{U_{t+1}(\mathbf{x}|\mathcal{D}_t)-L_{t+1}(\mathbf{x}|\mathcal{D}_t)} 
        & \text{\!\!\!\!\!if $y\! \in\! \Gamma_{t+1}^{\text{\tiny{LOCBO}}}(\mathbf{x}|\mathcal{D}_t)$,} \\
        \frac{\alpha \cdot p(y|\mathbf{x},\mathcal{D}_t)}{\lambda_{t+1}(\mathbf{x})} 
        & \text{\!\!\!\!\!otherwise}.
    \end{cases}
\end{align}
As shown in Figure \ref{fig: CBO illustration}, in a manner similar to CBO, the structure of the calibrated likelihood in \eqref{eq:locbo_likelihood} is motivated by the coverage properties of the prediction set $\Gamma_{t+1}^{\text{\tiny{LOCBO}}}(\mathbf{x}|\mathcal{D}_t) $. In fact, the calibrated likelihood  $p^{\text{\tiny{LOCBO}}}_\alpha(y|\mathbf{x},\mathcal{D}_t)$ in \eqref{eq:locbo_likelihood} assigns a  probability $1-\alpha$ to the interval $\Gamma_{t+1}^{\text{\tiny{LOCBO}}}(\mathbf{x}|\mathcal{D}_t) $ while being proportional to the GP likelihood $p(y|\mathbf{x},\mathcal{D}_t)$ for values outside the predicted region. Note that the normalization by $\lambda_{t+1}(\mathbf{x})$ in \eqref{eq:locbo_likelihood} ensures that the calibrated likelihood \eqref{eq:locbo_likelihood} is a valid density that integrates to 1.
\vspace{-8pt}
\subsection{Calibrated Posterior via Denoising}
\label{sec:posterior_calibration_locbo}
Based on the current GP posterior $p(f(\mathbf{x})|\mathcal{D}_t)$ and a likelihood $p^{\text{\tiny{LOCBO}}}_\alpha(y|\mathbf{x},\mathcal{D}_t)$ calibrated via localized CP to satisfy a target coverage level $\alpha\in[0,1]$ \cite{zecchin2024localized}, the posterior is obtained by using the denoising operation
\begin{align}	p^{\text{\tiny{LOCBO}}}_\alpha(f(\mathbf{x})|\mathcal{D}_t)\!=\!\!\!\int_{y'\in \mathcal{Y}}\!\!\!\!\!\! p(f(\mathbf{x})|\mathcal{D}_t\! \cup \! \{(\mathbf{x},y')\})p^{\text{\tiny{LOCBO}}}_\alpha(y'|\mathbf{x},\mathcal{D}_t)dy'\!.
 \label{eq:LOCBO_decomp}
\end{align}
The value of the posterior $p^{\text{\tiny{LOCBO}}}_\alpha(f(\mathbf{x})|\mathcal{D}_t)$ in \eqref{eq:LOCBO_decomp} is computed by averaging the value of the GP posterior $p(f(\mathbf{x})|\mathcal{D}_t\cup\{(\mathbf{x},y')\})$  in \eqref{eq: GP} for an observation $y'$ over the calibrated likelihood $p^{\text{\tiny{LOCBO}}}_\alpha(y|\mathbf{x},\mathcal{D}_t)$. Thus, the estimate \eqref{eq:LOCBO_decomp}  accounts for the knowledge extracted from the calibration process by localized online CP and for the model-based information encoded by the GP prior.

As shown in Appendix \ref{appendix:LOCBO_decomp}, the integral in \eqref{eq:LOCBO_decomp} can be evaluated in closed form based on the calibrated likelihood $p^{\text{\tiny{LOCBO}}}_\alpha(y|\mathbf{x},\mathcal{D}_t)$ in  \eqref{eq:locbo_likelihood}. Furthermore, based on LOCBO's posterior \eqref{eq:LOCBO_decomp}, the acquisition function \eqref{eq:acqusition function} can be evaluated as
\begin{align}
    \label{eq:locbo_utility}
    a^{\text{\tiny{LOCBO}}}(\mathbf{x}|\mathcal{D}_t) &= \int  u(\mathbf{x}, f(\mathbf{x}), \mathcal{D}_t)p^{\text{\tiny{LOCBO}}}_\alpha(f(\mathbf{x})|\mathcal{D}_t)\mathrm{d}f,
\end{align}
and estimated by using standard Monte Carlo integration. The next query point  $\mathbf{x}_{t+1}$  is finally selected as the maximizer of the estimated expected utility:
\begin{align}
\label{eq:locbo_iterate}
\mathbf{x}_{t+1}=\argmax_{\mathbf{x}}a^{\text{\tiny{LOCBO}}}(\mathbf{x}|\mathcal{D}_t).
\end{align}

\subsection{Theoretical Guarantees}
\label{sec:theoretical_guarantees_locbo}
In this section, we present the theoretical guarantees of LOCBO. We first relate the properties of the observation noise $\xi(\mathbf{x}) $ in \eqref{eq:noise_modelling} to the coverage properties of the prediction set $\Gamma_{t}^{\text{\tiny{LOCBO}}}(\mathbf{x}|\mathcal{D}_{t-1})$ with respect to the unknown function $f(\mathbf{x})$. Building on this intermediate result, we then establish a lower bound on the utility of LOCBO’s iterates \eqref{eq:locbo_iterate} in terms of the estimated expected utility \eqref{eq:locbo_utility}. This result certifies the effectiveness of the calibrated acquisition function \eqref{eq:locbo_utility} as a guiding principle for selecting future queries.

As detailed in Algorithm \ref{alg:LOCBO}, LOCBO applies localized online CP to produce a sequence of prediction sets $\{\Gamma_{t}^{\text{\tiny{LOCBO}}}(\mathbf{x}|\mathcal{D}_{t-1})\}_{t\geq 1}$ based on the queries and observations $\{(\mathbf{x}_t,y_t)\}_{t\geq 1}$. As stated in the following lemma, the calibration properties of localized CP ensure that LOCBO’s prediction sets $\{\Gamma_{t}^{\text{\tiny{LOCBO}}}(\mathbf{x}|\mathcal{D}_{t-1})\}_{t \geq 1}$ satisfy a long-run coverage guarantee.

\begin{assumption}[Stationary and bounded kernel]
		\label{ass:lower_bound_k}
		The kernel function is stationary, i.e., $k_g(x,x')=\tilde{k}_g(\lVert x-x'\rVert)$, for some non-negative function $\tilde{k}(\cdot)$, which is $\rho$-Lipschitz for some $\rho>0$, upper bounded by $\kappa< \infty$, and coercive, i.e., $\lim_{z\to \infty}\tilde{k}(z)=0$.
\end{assumption}

\begin{lemma}[Long-run coverage of the noisy observation \cite{zecchin2024localized}]
    \label{lemma:long-run_noisy}
        Fix a user-defined target miscoverage level $\alpha\in[0,1]$. Under Assumption \ref{ass:lower_bound_k}, for any hyperparameter $\lambda>0$ and any learning rate sequence $\eta_t=\eta_1 t^{-1/2}< 1/\lambda$ with $\eta_1>0$, given any query-observation sequence $\{(\mathbf{x}_t,y_t)\}^T_{t=1}$ with bounded input $\lVert \mathbf{x}_t\rVert\leq D<\infty$, LOCBO's set predictors in (\ref{eq:locbo_prediction_set}) satisfy the deterministic coverage condition
        \begin{align}
        \label{eq:mistake_bound_th}
			\frac{1}{T}\sum^T_{t=1} \mathbbm{1}\{y_t\notin \Gamma_{t}^{\textnormal{\tiny{LOCBO}}}(\mathbf{x}_t|\mathcal{D}_{t-1}) \}\leq  \alpha+ \frac{\beta}{\sqrt{T}} +{\color{black}\kappa},
		\end{align}
        with $\beta=\frac{2}{\eta_1 }+\frac{4 \sqrt{\rho{\color{black}\kappa D}}}{\eta_1\lambda}+2(2\kappa+1)$.
\end{lemma}

Lemma \ref{lemma:long-run_noisy} provides a bound on the number of miscoverage errors that LOCBO’s prediction sets incur with respect to the actual noisy observations $y_t$ over a time horizon of $T$ steps. Unlike the online CP guarantee \eqref{eq:locp_guarantee}, which states that the long-run average number of errors converges to the target miscoverage rate $\alpha$, localized CP ensures that this quantity converges to a neighborhood of $\alpha$ with a radius of $\kappa$, which is proportional to the degree of localization of the kernel. Notably, in the absence of localization ($\kappa=0$), this radius can be shown to be zero, thereby recovering the OCBO guarantees \eqref{eq: calibration condtion} as a special case.

Like the OCBO's guarantee \eqref{eq: calibration condtion}, Lemma \ref{lemma:long-run_noisy} guarantees calibration with respect to the noisy observation $y_t$. However, this does not imply coverage of the actual optimization objective $f(\mathbf{x}_t)$ in the presence of noise, which is assumed in the observation \eqref{eq:noise_modelling}. To extend the results to a coverage guarantee for the objective $f(\mathbf{x}_t)$, we make the following basic assumption about the observation noise $\xi(\cdot)$.

Define the quantity
\begin{align}
    \label{eq:noise_imbalance} 
    b_{\xi}:=\inf_{x\in\mathcal{X}}\left[\min\left(\Pr[\xi(x)\geq0],\Pr[\xi(x)\leq0]\right)\right],
\end{align}
which captures the minimum probability that the noise $\xi(x)$ is positive or negative. Thus, the value $b_{\xi}$ quantifies the level of symmetry in the noise distribution across the input space $\mathcal{X}$ \cite{guille2024conformal}. A higher value of the scalar $b_{\xi}$ indicates a more balanced noise distribution, while a lower value suggests a greater imbalance. For instance, if the noise has a median equal to 0, we have $b_{\xi}=0.5$, representing the most balanced noise. In contrast, a zero value for the scalar $b_{\xi}$, i.e., $b_{\xi}=0$, corresponds to an observation noise case that is positive or negative with probability 1. 

\begin{assumption}
    \label{ass:balanced_noise}
    The observation noise is not maximally imbalanced, i.e., 
    \begin{align}
        b_\xi>0.
    \end{align}
\end{assumption}

As stated in the following lemma, for strictly positive values of the symmetry parameter $b_{\xi}$, it is possible to obtain a coverage guarantee with respect to the objective function $f(\cdot)$.

\begin{lemma}[Long-run coverage of the objective function $f(\cdot)$ ]
    \label{lemma:long-run_obj}
        Fix a user-defined target miscoverage level $\alpha\in[0,1]$. Under Assumptions \ref{ass:lower_bound_k} and \ref{ass:balanced_noise}, for any sequence of queries $\{\mathbf{x}_t\}^T_{t=1}$, the set predictors produced by LOCBO in (\ref{eq:locbo_prediction_set}) satisfy the average coverage condition
        \begin{align}
        \label{eq:mistake_bound_th_noiseless}
			\frac{1}{T}\sum^T_{t=1} \Pr[f(\mathbf{x}_t)\notin \Gamma_{t}^{\textnormal{\tiny{LOCBO}}}(\mathbf{x}_t|\mathcal{D}_{t-1}) ]\leq  \frac{1}{b_{\xi}}\left(\alpha+ \frac{\beta}{\sqrt{T}} +{\color{black}\kappa}\right),
		\end{align}
        where the probability is taken over the observation noise sequence $\{\xi(\mathbf{x}_t)\}^T_{t=1}$.
\end{lemma}

Interestingly, the coverage guarantee in Lemma \ref{lemma:long-run_obj} holds with respect to the objective function $f(\mathbf{x}_t)$, which is never observed during the calibration process. For this reason, however, the guarantee is inevitably  probabilistic rather than deterministic, as in Lemma \ref{lemma:long-run_noisy}. Specifically, inequality \eqref{eq:mistake_bound_th_noiseless} provides a bound on the long-run average probability of the predicted sets \eqref{eq:locbo_prediction_set} not covering the true objective $f(\mathbf{x}_t)$. The bound \eqref{eq:mistake_bound_th_noiseless}, aside from algorithmic constants, also depends on the noise symmetry $b_{\xi}$, which appears as the multiplicative factor $1/b_{\xi}$ in the achievable miscoverage error.

Based on the calibrated sets $\{\Gamma_{t}^{\text{\tiny{LOCBO}}}(\mathbf{x}|\mathcal{D}_{t-1})\}_{t \geq 1}$, LOCBO computes the calibrated posterior \eqref{eq:LOCBO_decomp}, which is then used to estimate the expected utility \eqref{eq:locbo_utility}. We now show that, thanks to the coverage guarantee in Lemma \ref{lemma:long-run_obj}, the expected utility \eqref{eq:locbo_utility}, which is estimated by using the LOCBO's posterior \eqref{eq:LOCBO_decomp}, provides a lower bound on the actual utility of the iterates. To this end, we make the following assumptions regarding the utility function and the GP posterior \eqref{eq: GP}.

\begin{assumption}[Increasing and bounded positive utility]
		\label{ass:decreasing_utility}
		The utility function $u(\mathbf{x}, f(\mathbf{x}), \mathcal{D}_t)$ is bounded i.e., there exists $C<\infty$ such that $u(\cdot, \cdot, \cdot) \leq C$, and increasing in the value of the objective function $f(\mathbf{x})$ for any fixed $\mathbf{x}$ and $\mathcal{D}_t$.
\end{assumption}
Assumption \ref{ass:decreasing_utility} states that the utility of querying an input $\mathbf{x}$ on the basis of evidence $\mathcal{D}_t$, should be larger when the function $f(\mathbf{x})$ is larger. Such an assumption is coherent with the goal of finding the value $\mathbf{x}$ that maximizes the optimization objective $f(\cdot)$.
\begin{assumption}[Lower bound on GP posterior conservativeness]
		\label{ass:conservative_posterior}
		There exists a constant $0<\epsilon\leq 1$, such that the GP posterior $p(f(\mathbf{x})|\mathcal{D}_t\cup\{(\mathbf{x},y)\})$ defined in \eqref{eq: GP} satisfies the inequality
        \begin{align}
            \int^{y}_{-\infty}p(y'|\mathcal{D}_t\cup\{(\mathbf{x},y)\})dy'\geq \epsilon.
        \end{align}
        for all observation sequences $\mathcal{D}_t$ and pair $\{(\mathbf{x},y) \}$.
\end{assumption}
Assumption \ref{ass:conservative_posterior} can be interpreted as a bound on the probability that the GP posterior $p(f(\mathbf{x})|\mathcal{D}_t\cup\{(\mathbf{x},y)\})$ assigns to the event $\{f(\mathbf{x}) < y\}$ based on the observations $\mathcal{D}_t\cup\{(\mathbf{x},y)\}$. As such, Assumption \ref{ass:conservative_posterior} can be interpreted as a lower bound on the conservativeness of the denoising distribution, with larger values of $\epsilon$ leading to more conservative GP posteriors.

\begin{theorem}
    \label{th:utility_guarantee}
	For any $\alpha\in[0,1]$, under Assumptions \ref{ass:lower_bound_k}-\ref {ass:conservative_posterior}, and setting, without loss of generality, $C=0$ in Assumption \ref{ass:decreasing_utility}, the iterates $\{\mathbf{x}_{t}\}^T_{t=1}$ produced by LOCBO via \eqref{eq:locbo_iterate} satisfy the inequality
	\begin{align}
        \label{eq:locbo_utility_guarantee}
		\frac{1}{T}\sum^T_{t=1}\Pr\left[u(\mathbf{x}_{t},f(\mathbf{x}_{t}),\mathcal{D}_{t-1})\geq\frac{2 a^{\textnormal{\tiny{LOCBO}}}(\mathbf{x}_{t}|\mathcal{D}_{t-1})}{\alpha\epsilon}\right] \nonumber \\ \geq 1 - \frac{1}{b_{\xi}}\left(\alpha+ \frac{\beta}{\sqrt{T}} +{\color{black}\kappa}\right).
	\end{align}
\end{theorem}
Theorem \ref{th:utility_guarantee} provides a probabilistic lower bound on the utility of the iterates produced by LOCBO. The lower bound is proportional to the expected utility $a^{\text{\tiny{LOCBO}}}(\mathbf{x}_{t}|\mathcal{D}_{t-1})$ estimated by LOCBO, indicating that LOCBO utility estimate $a^{\text{\tiny{LOCBO}}}(\mathbf{x}_{t}|\mathcal{D}_{t-1})$
provides a meaningful objective for selecting query points. Indeed, by \eqref{eq:locbo_utility_guarantee}, a large value of the utility $a^{\text{\tiny{LOCBO}}}(\mathbf{x}_{t}|\mathcal{D}_{t-1})$ also indicates a large value of the true utility $u(\mathbf{x}_{t},f(\mathbf{x}_{t}),\mathcal{D}_{t-1})$. 

Furthermore, the theorem highlights the detrimental effect of noise $b_{\xi}$, and it provides insight into the choice of the miscoverage level $\alpha$.  Specifically, in order to maintain a given multiplicative gap between the actual and estimated utilities in the left-hand side of \eqref{eq:locbo_utility_guarantee}, the target miscoverage rate $\alpha$ must be ideally selected to be inversely proportional to the conservativeness of the GP posterior as measured by the parameter $\epsilon$. In practice, as we will discuss in the next section, this result implies that a suitable choice of the target coverage $\alpha$ can compensate for errors due to the misspecification of the GP prior.
\begin{algorithm}[t]
\caption{Localized Online CP-based Bayesian Optimization (LOCBO)}
\begin{algorithmic}[1]
\label{alg:LOCBO}
\REQUIRE GP prior $p(f(\mathbf{x}))$, target miscoverage level $\alpha\in[0,1]$, kernel $k_g(\cdot,\cdot)$, hyperparameter $\lambda>0$, learning rate sequence $\eta_t=\eta_1 t^{-1/2}< 1/\lambda$ with $\eta_1>0$ and $\mathcal{D}_0=\emptyset$.
\FOR{$t=1,\ldots, T_{\text{max}}$}
    \STATE Compute the calibrated likelihood $p^{\text{\tiny{LOCBO}}}_\alpha(y|\mathbf{x},\mathcal{D}_{t-1})$ as in \eqref{eq:locbo_likelihood}
    \STATE Obtain the posterior $p_\alpha^{\text{\tiny{LOCBO}}}(f(\mathbf{x})|\mathcal{D}_{t-1})$ via the denoising operation \eqref{eq:LOCBO_decomp}

    \STATE Select the query $\mathbf{x}_{t}$ via \eqref{eq:locbo_iterate}
     \STATE Observe $y_{t}$ for query $\mathbf{x}_{t}$ and update the evidence $\mathcal{D}_{t}$ = $\mathcal{D}_{t-1} \cup \{(\mathbf{x}_{t},y_{t})\}$
     \STATE Update the GP posterior distribution $p(f(\mathbf{x})|\mathcal{D}_t)$ as in \eqref{eq: GP}
     \STATE Compute the updated localized CP threshold $\lambda_{t+1}(\cdot)$ via \eqref{eq:locp_update_rule_const}.
\ENDFOR
\ENSURE Return $\hat{\mathbf{x}}=\mathbf{x}_{t^*}$ for time $t^* = \argmax_{t=1,\dots,T_{\text{max}}} y_{t}$\\
\end{algorithmic}
\end{algorithm}
\section{Experimental Results}
\label{sec:Exp_results}
In this section, we evaluate the performance of the proposed LOCBO scheme against the state-of-the-art CP-based BO schemes reviewed in Section \ref{sec:Prior Art}. Firstly, we examine the prediction intervals obtained by LOCBO under different localization levels. Next, we consider a BO problem with a synthetic 2-dimensional Ackley function \cite{surjanovic2013virtual}. For this objective, we examine both homoscedastic and heteroscedastic observation noise. Finally, we study a real-world engineering problem related to radio resource management in unmanned aerial vehicle (UAV) networks \cite{benzaghta2023designing}. The code for the experiments is available at https://github.com/davinci003/LOCBO.
\vspace{-10pt}
\subsection{Simulation Setup}
In all  experiments, we adopt a GP prior with a Mat\'ern-5/2 kernel  \cite{snoek2012practical}
\begin{align}
\label{eq:matern_kernel}
k(\mathbf{x}, \mathbf{x}')\!\! =\!\! \left(\! 1\!\! +\!\! \frac{\sqrt{5} \lVert\mathbf{x}-\mathbf{x}'\rVert}{\gamma}\!\! +\!\! \frac{5 \lVert\mathbf{x}-\mathbf{x}'\rVert^2}{3 \gamma^2}\!\! \right)\!\exp\!\! \left(\!\! -\frac{\sqrt{5} \lVert\mathbf{x}-\mathbf{x}'\rVert}{\gamma}\!\! \right).
\end{align}
The length scale $\gamma$, together with the observation noise variance $\sigma^2$ in \eqref{eq: observations}, are updated via maximum likelihood during optimization \cite{frazier2018tutorial}. The expected improvement utility function is used for the acquisition function \eqref{eq:acqusition function}.

We evaluate the performance of the following BO schemes:
\begin{itemize}[leftmargin=11pt]

\item CBO, reviewed in Section \ref{sec:CBO}, with likelihood ratios obtained via a parametric estimator based on samples generated from multiple stochastic gradient Langevin dynamics (SGLD) chains \cite{welling2011bayesian} as in \cite{stanton2023bayesian};
\item OCBO, reviewed in Section \ref{sec:OCBO}; 
\item OCBO-L, a variant of OCBO obtained by using localized online CP \cite{zecchin2024localized} in lieu of online CP to optimize the recalibration function $R_t(\alpha)$.
\item The proposed LOCBO algorithm. For both OCBO-L and LOCBO, the RBF kernel 
\begin{align}
    k_{g}(\mathbf{x},\mathbf{x}’)= \kappa\exp{\left(\frac{-\lVert\mathbf{x}-\mathbf{x}’\rVert^2}{l^2}\right)}
    \label{eq:LOCBO_RBF_kernel}
\end{align}
is adopted to implement the localized online CP update \eqref{eq:locp_update_rule_const}.
The scaling parameter $\kappa$ and the length scale $l$ in \eqref{eq:LOCBO_RBF_kernel} control the localization properties of localized online CP \cite{zecchin2024localized}. We fix $\kappa$, while varying the length scale $l$ to tune the localization of LOCBO, with a smaller $l$ corresponding to a more localized kernel. We will also consider the degenerate case $l=\infty$, which yields a non-localized version of LOCBO.
\end{itemize}
\begin{figure}
     \centering
     \begin{subfigure}[b]{0.5\columnwidth}
         \centering
         \includegraphics[width=\columnwidth]{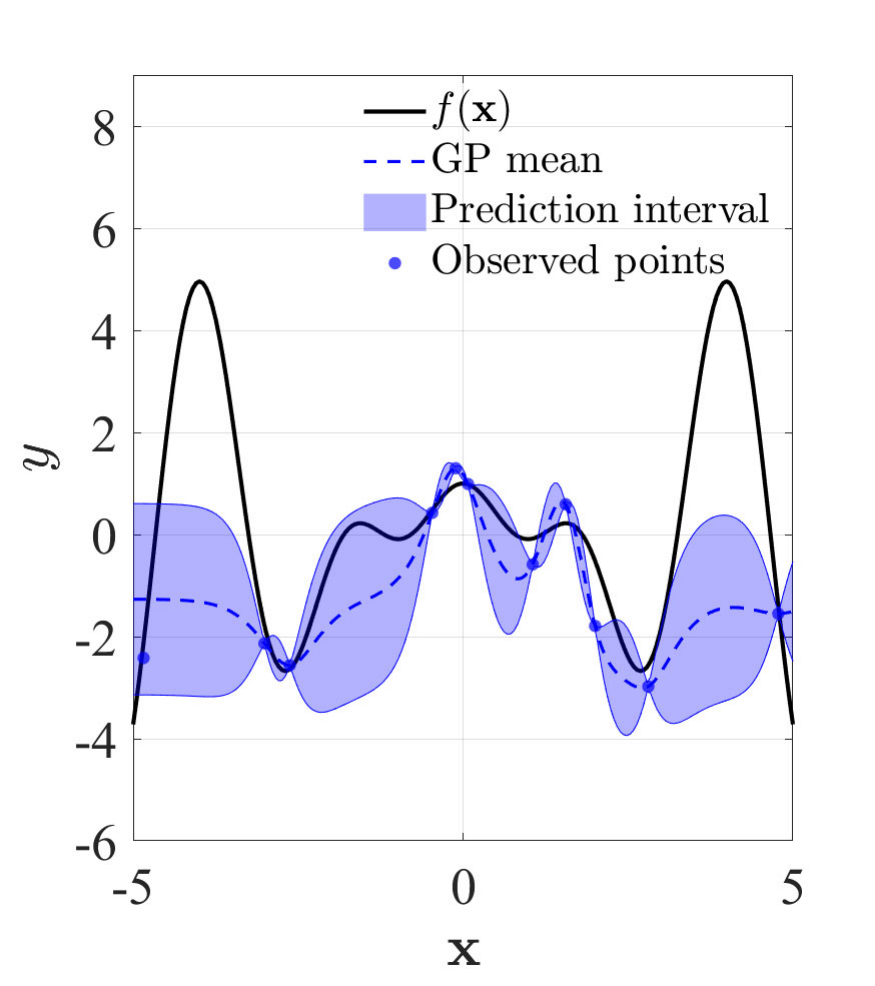}
         \caption{}
         \label{fig: Ackely pure}
     \end{subfigure}
     \hfill
     \hspace{-11pt}
     \begin{subfigure}[b]{0.5\columnwidth}
         \centering
         \includegraphics[width=\columnwidth]{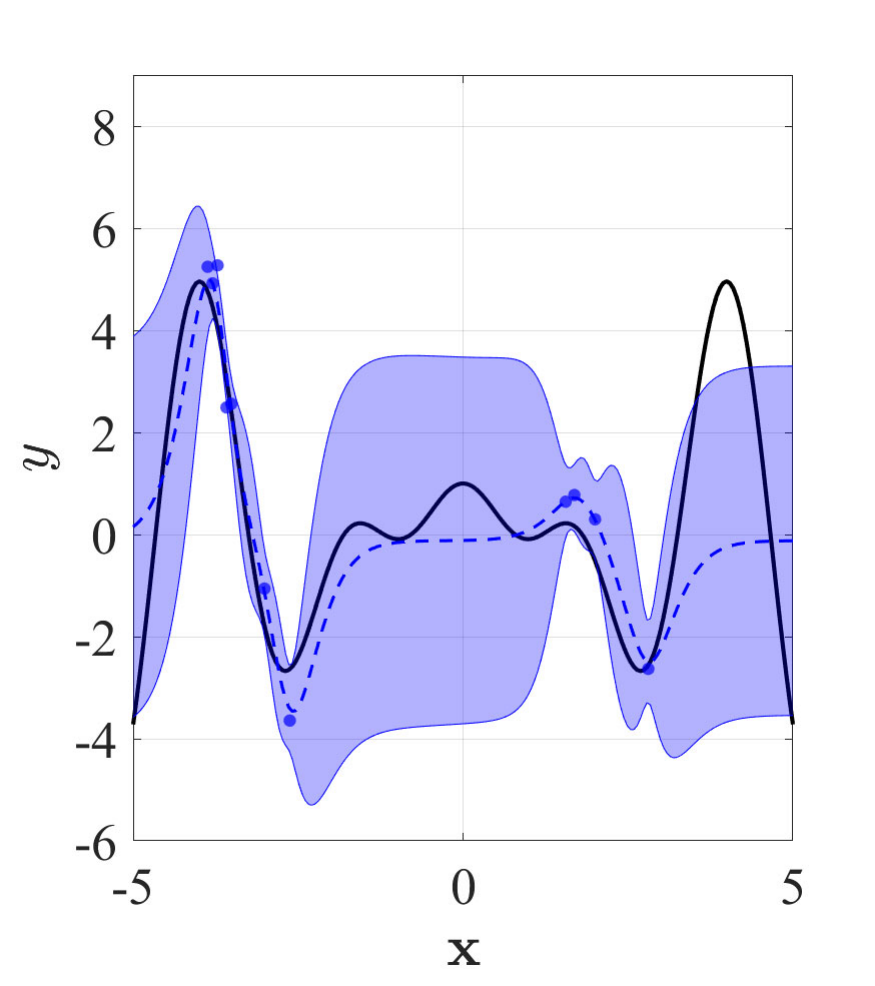}
         \caption{}
         \label{fig: Ackley noise}
     \end{subfigure}
     \hfill
        \caption{Prediction intervals obtained at the end of the optimization process for LOCBO with (a) $\kappa = 0$ and (b) $\kappa = 2$ on a synthetic function (black line). With a larger localization, i.e., for a larger $\kappa$, LOCBO can focus the reduction of uncertainty on the portions of the optimization domain that are closest to optimal values.}
        \vspace{-0.5cm}
    \label{fig:prediction_interval}
\end{figure}
 We emphasize that, while this work proposes LOCBO, the scheme OCBO-L has also not been considered before.
 
All online-CP-based methods use the same learning rate $\eta_t=\eta_0t^{-w}$, with the initial learning rate $\eta_0 = 5\times10^{-3}$ in Sections \ref{sec: experiment prediciton interval}, \ref{sec: experiment synthetic objective function}, and \ref{sec: experiment RRM}. The decay rate $w$ is set to $5\times10^{-2}$ in Sections \ref{sec: experiment prediciton interval} and \ref{sec: experiment synthetic objective function}, and to $5\times10^{-3}$ in Section \ref{sec: experiment RRM}. For OCBO-L and LOCBO, the RBF kernel's length scale $l$ is set to $5$ in Section \ref{sec: experiment synthetic objective function} and to $1/3$ in Section \ref{sec: experiment RRM}. The scaling parameter $\kappa$ of OCBO-L and LOCBO is set to $5$ and $4$ in Section \ref{sec: experiment synthetic objective function} and to $2$ in Section \ref{sec: experiment RRM}. Lastly, the regularization hyperparameter $\lambda$ is set to $4\times 10^{-3}$ in Sections \ref{sec: experiment prediciton interval} and \ref{sec: experiment synthetic objective function}, and to $1\times 10^{-4}$  in Section \ref{sec: experiment RRM}.

\subsection{Prediction Interval for LOCBO}
\label{sec: experiment prediciton interval}
Figure 4 illustrates the impact of localization on the prediction interval. Specifically, we consider the maximization of the one-dimensional function
\begin{align}
    f(\mathbf{x}) = \mathbf{x}\sin{(2\mathbf{x})}+\cos{(\pi \mathbf{x})} \tag{41}
    \label{eq:synthetic_prediction_interval}
\end{align}
over the domain $\mathbf{x} \in [-5,5]$. The observation noise is modeled as $\xi(\mathbf{x}) \sim \mathcal{N}(0, \sigma^2_\xi(\mathbf{x}))$ with input-dependent variance $\sigma^2_\xi(\mathbf{x}) = (\lVert \mathbf{x} \rVert + 1) / 10$. In the conventional case $\kappa=0$, the prediction intervals tend to be uniform across the optimization domain. In contrast, a larger localization, obtained with $\kappa=2$, enhances exploration near the global optimum, reducing uncertainty in the most relevant part of the optimization domain.

\subsection{Synthetic Objective Function}
\label{sec: experiment synthetic objective function}
In this section, we consider the maximization over the domain $\mathbf{x}\in[-10, 10]^2$ of the 2-dimensional Ackley function
\begin{align}
    f(\mathbf{x}) =& a\exp{\left(-b\sqrt{\frac{1}{2}\sum_{i=1}^2 x_i^2}\right)}  + \exp{\left(\frac{1}{2}\sum_{i=1}^2\cos{(cx_i)}\right)}\nonumber \\& - a - \exp{(1)}
    \label{eq:2D_ackley}
\end{align}
with parameters $a = 20$, $b=0.2$,  and $c= 2\pi$  \cite{surjanovic2013virtual}. 
The GP model is initialized using five random, uniformly distributed input values, and we consider an optimization horizon of $T=50$ with CBO, LOCBO ($l =\infty$), and LOCBO targeting a miscoverage level $\alpha=0.2$. Performance is evaluated via the simple regret
\begin{align}
\label{eq:simple_regret}
r_{T} = \max_{\mathbf{x}\in \mathcal{X}}f(\mathbf{x}) - f(\mathbf{x}_{t^*}),
\end{align}
\begin{figure}
     \centering
     \begin{subfigure}[b]{0.5\columnwidth}
         \centering
         \includegraphics[width=\columnwidth]{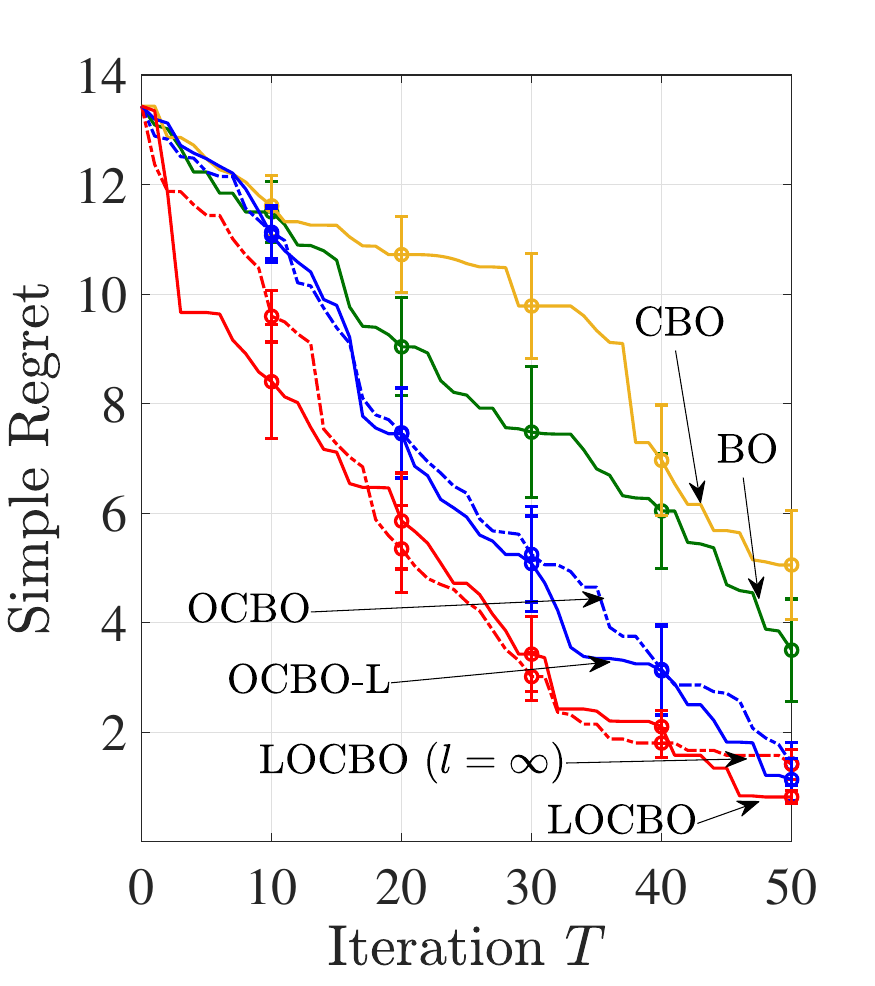}
         \caption{}
         \label{fig: Ackely pure}
     \end{subfigure}
     \hfill
     \hspace{-11pt}
     \begin{subfigure}[b]{0.5\columnwidth}
         \centering
         \includegraphics[width=\columnwidth]{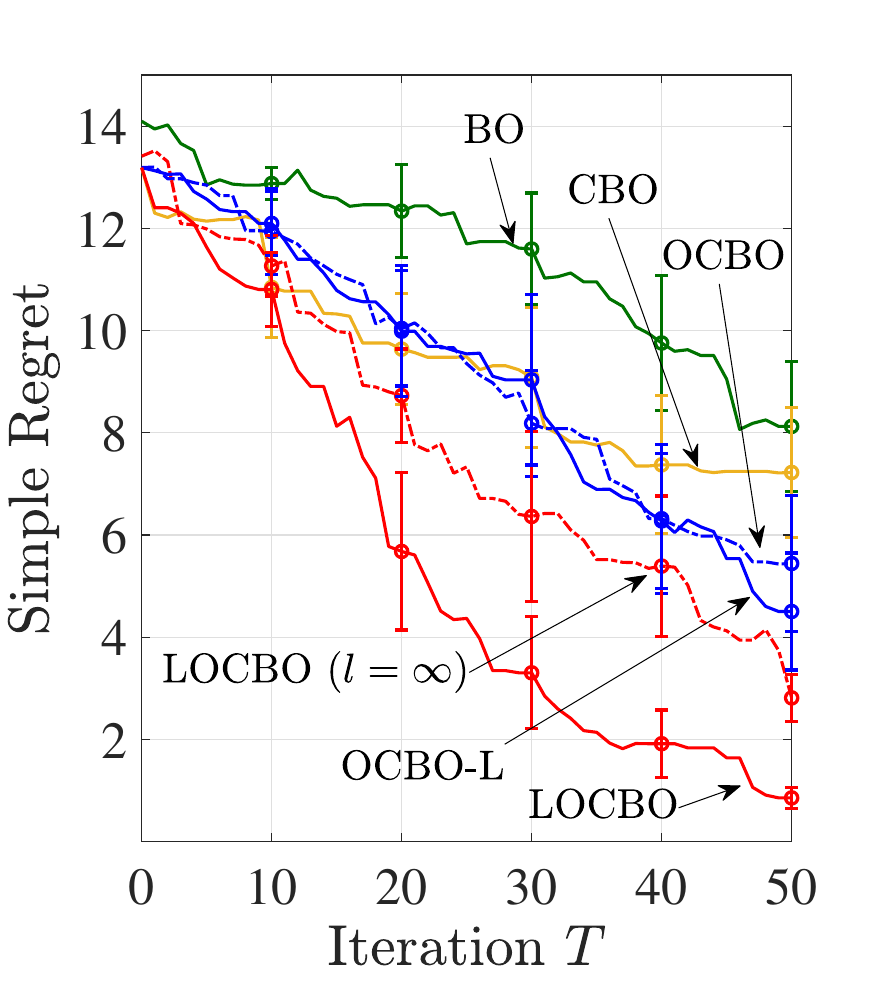}
         \caption{}
         \label{fig: Ackley noise}
     \end{subfigure}
     \hfill
        \caption{(a) Simple regret \eqref{eq:simple_regret} as a function of the optimization horizon $T$ for the Ackley 2D function without observation noise. (b) Simple regret \eqref{eq:simple_regret} as a function of the optimization horizon $T$ for the Ackley 2D function with observation noise. All figures show mean and 70\% confidence interval. BO, CBO, OCBO, LOCBO ($l=\infty$), OCBO-L, and LOCBO correspond to green, yellow, blue dashed, blue solid, red dashed, and red solid line, respectively.}
        \vspace{-0.5cm}
\end{figure}
where $t^* = \argmax_{t=1,\dots,T} y_{t}$, which measures the suboptimality gap of the best solution $\mathbf{x}_{t^*}$ available at time $T$\cite{wang2017max}.

In Figure \ref{fig: Ackely pure}, for reference, we consider a setting in which the observation noise is absent, i.e., $\xi(\mathbf{x})=0$ for all $\mathbf{x}$. We report the simple regret averaged over 5 trials as a function of the optimization round $T$. Generally, CP calibration improves BO performance, resulting in lower regret across the entire optimization horizon. The only exception is CBO, which was found not to improve over BO for this example. The proposed LOCBO scheme outperforms all other methods, although OCBO offers similar performance in the absence of observation noise. As another observation, comparing LOCBO and OCBO-L to their respective non-localized variants, LOCBO with length scale $l=\infty$ and OCBO, it is concluded that localization provides limited benefits in this setup. 

We now consider a more challenging scenario in which observations are affected by heteroscedastic Gaussian noise. In particular we model the observation noise as $\xi(\mathbf{x}) \sim \mathcal{N}(0, \sigma^2_\xi(\mathbf{x}))$, where the variance $\sigma^2_\xi(\mathbf{x}) = (\lVert \mathbf{x} \rVert + 10) / 20$ increases linearly with the input magnitude $\lVert \mathbf{x} \rVert$. Standard BO assumes uniform variance across all inputs, causing the likelihood to be misspecified. Figure \ref{fig: Ackley noise} shows the simple regret averaged over 7 experiment runs as a function of the number of iterations,  $T$. For this more complex problem, LOCBO, which incorporates a carefully designed denoising step, achieves significantly better performance than all other schemes. In contrast, OCBO, which assumes noiseless observations, shows a significant performance degradation in the noisy observation setting at hand. Additionally, the benefit of localization becomes apparent in the presence of heteroscedastic observation noise. Notably, LOCBO achieves a final simple regret less than half of that obtained without localization, i.e., with $l=\infty$.
\begin{figure}
     \centering
     \begin{subfigure}[b]{0.5\columnwidth}
         \centering
         \includegraphics[width=\columnwidth]{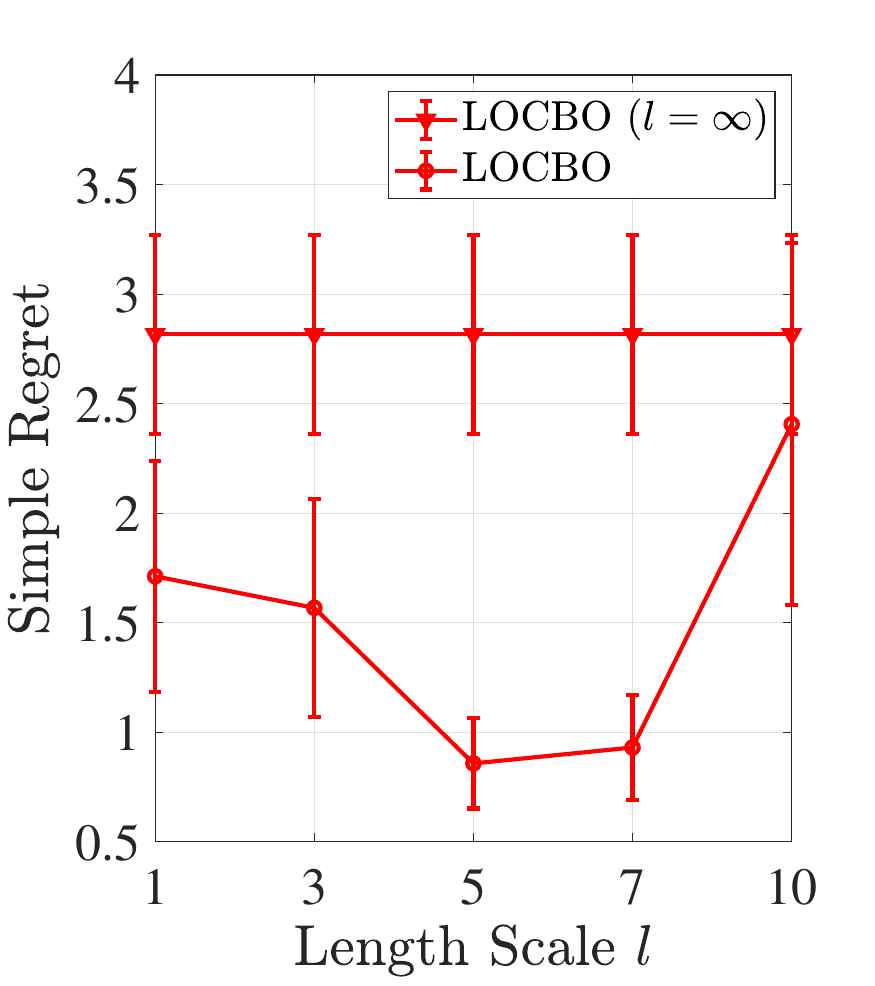}
         \caption{}
         \label{fig: Ackley noise LOCBO per bandwidth}
     \end{subfigure}
     \hfill
     \hspace{-11pt}
     \begin{subfigure}[b]{0.5\columnwidth}
         \centering
         \includegraphics[width=\columnwidth]{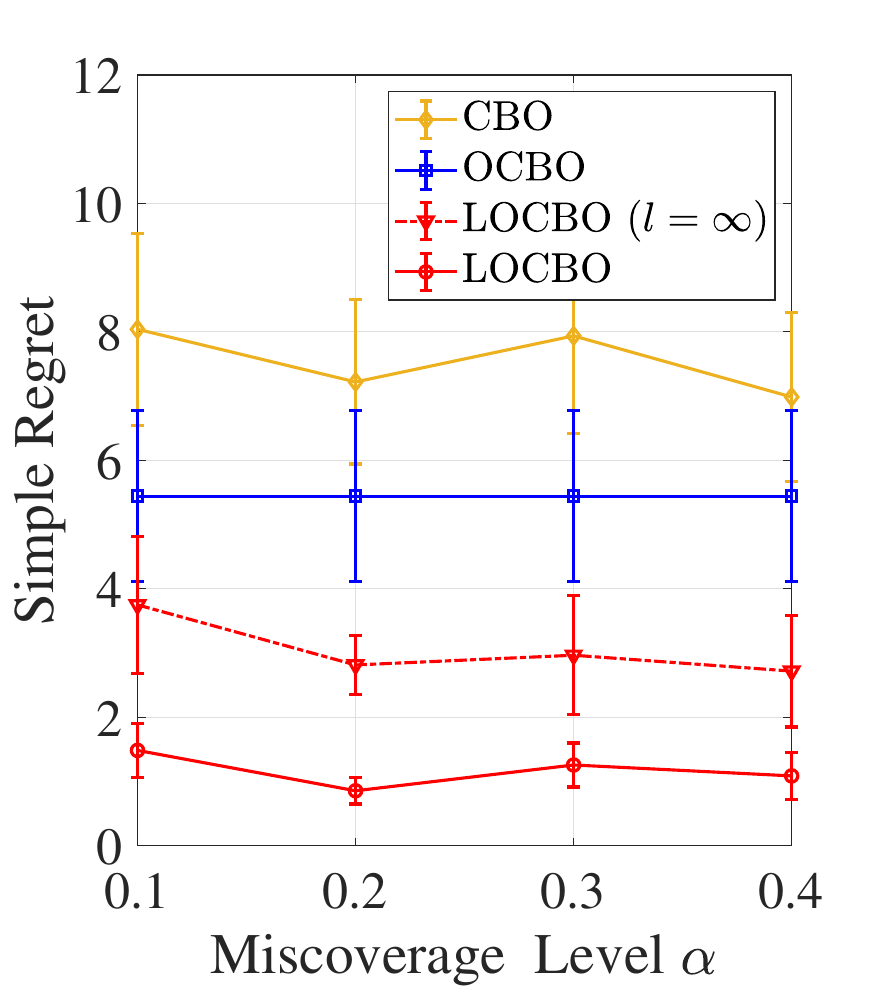}
         \caption{}
         \label{fig: Ackley per alpha}
     \end{subfigure}
     \hfill
        \caption{(a) Simple regret \eqref{eq:simple_regret} as a function of the length scale $l$ for the Ackley 2D function with observation noise. 
    (b) Simple regret \eqref{eq:simple_regret} as a function of the target miscoverage level $\alpha$ for the Ackley 2D function with observation noise. All figures show mean and 70\% confidence interval.}
        \label{fig:three graphs}
        \vspace{-0.4cm}
\end{figure}

To further study the impact of localization on the performance of LOCBO, Figure \ref{fig: Ackley noise LOCBO per bandwidth} reports the simple regret obtained by LOCBO for different length scale values $l$ at the optimization horizon $T=50$. Finite values of the localization parameters yield better performance levels compared to non-localized LOCBO with $l=\infty$. However, the performance gain is not monotonic in the length scale $l$. The smallest simple regret is obtained for $l=5$, with smaller values of $l$ degrading the performance of LOCBO. 

In Figure \ref{fig: Ackley per alpha}, we examine the effect of varying the miscoverage level $\alpha$ on the performance of CP-based BO schemes. The simple regret, averaged over $7$ trials, is reported at different miscoverage levels $\alpha$ at the end of the optimization horizon $T=50$. Across all tested values of $\alpha$, LOCBO demonstrates superior performance in minimizing the simple regret compared to the baseline methods. Furthermore, all schemes prove to be quite robust to the choice of the miscoverage level $\alpha$. This suggests that the choice of $\alpha$ may not be as critical as the selection of the length scale $l$.

\begin{figure}[t]
\centering
\begin{subfigure}[t]{0.5\linewidth} 
    \centering
    \includegraphics[width=\linewidth]{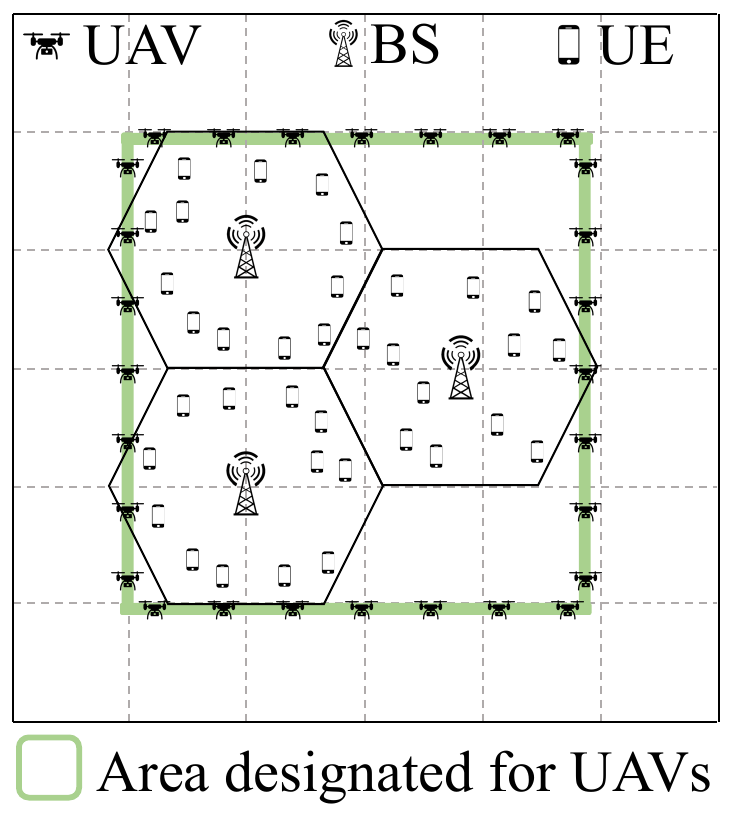} 
    \caption{}
    \label{fig:figure_cuav_deployment}
\end{subfigure}
\hfill 
\hspace{-10pt}
\begin{subfigure}[t]{0.5\linewidth}
    \centering
    \includegraphics[width=\linewidth]{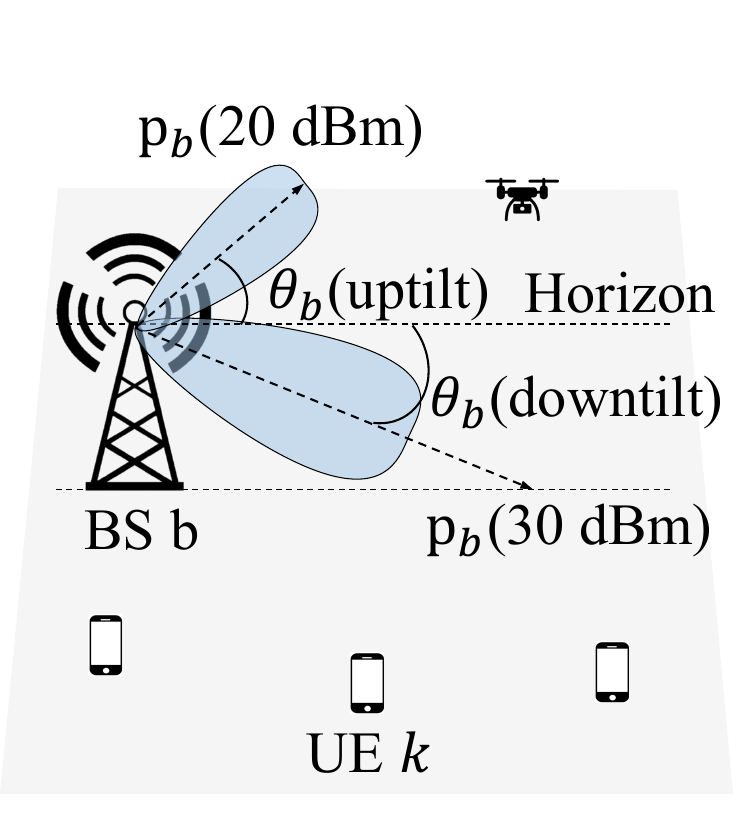} 
    \caption{}
    \label{fig:figure_cuav_antenna_tilt}
\end{subfigure}
\caption{(a) Network deployment for the radio resource management problem studied in Section \ref{sec: experiment RRM}. The network consists of $N_{\text{BS}}= 9$ BSs, $N_{\text{GU}} = 36$ GUs, and $N_{\text{UAV}} = 28$ UAVs. (b) Example of transmit power and antenna tilt for a BS $b$. The transmit power $p_b$ lies within the range of $[6\text{ dBm}, 46\text{ dBm}]$, and the antenna tilt $\mathbf{\theta}_b$ lies within the range of $[-90^{\circ}, 90^{\circ}]$.}
\label{fig:ccuav_deployment}
\vspace{-0.6cm}
\end{figure}

\subsection{Radio Resource Management}
\label{sec: experiment RRM}
We now examine the application of BO to the engineering problem of radio resource allocation in cellular-connected UAV networks. As detailed in \cite{benzaghta2023designing}, this problem requires the optimization of antenna tilt angles and of the transmission powers at multiple cellular base stations to improve the connectivity to UAVs. In particular, as illustrated in Figure \ref{fig:ccuav_deployment}, we consider a 3GPP compliant cellular network \cite{specification3gpp} with a hexagonal layout and comprising three cells, each with a 200-meter inter-cell distance. Each cell is served by $N_{\text{BS}}=3$ co-located base stations (BSs) with a height of 25 meters, each covering a 120° azimuth sector. We denote the set of BSs as $\mathcal{B}$. We assume that the BSs serve $N_{\text{GU}}$ ground users (GUs) and  $N_{\text{UAV}}$ UAVs, which we respectively denote as $\mathcal{G}$, and $\mathcal{U}$. The GUs are uniformly distributed within the cells, while the UAVs are evenly arranged within four designated rectangular regions.

The objective is to maximize the average capacity $f(\mathbf{x})$ for both GUs and UAVs over the vector $\mathbf{x}=[\mathbf{p},\mathbf{\Theta}]$, which includes the base stations transmit powers $\mathbf{p} \in \mathbbm{R}^{N_{\text{BS}}\times 1}$ and the antenna tilts  $\mathbf{\Theta} \in \mathbbm{R}^{N_{\text{BS}}\times 1} $. Define as $h$ the collection of all fading channels $\{h_{b,k}\}$ between BS $b$ and user $k$. The channels follow Rayleigh fading, and are complex Gaussian with zero mean and power equal to $1$. A user may be a GU or an UAV. Also, denote as $b_k$ the BS corresponding to the cell sector in which user $k$ is associated. For each BS $b$, the goal is optimizing transmit power $p_b$ and antenna tilt $\theta_b$.

The achievable rate of user $k$ is given by
\begin{align}
R_k (h)\!\! =\!\! \log_2\! \left(\!1\! +\! \frac{p_{b_k}\! \cdot\! g_{b_k,k}(\theta_{b_k})\! \cdot\! |h_{b_k,k}|^2}{\sum_{b \in \mathcal{B} \setminus \{b_k\}} p_b\! \cdot\! g_{b,k}(\theta_{b})\! \cdot\! |h_{b,k}|^2\! +\! \sigma^2_n}\!\right)\!,
\label{eq:CUAV_individual_objective_func}
\end{align}
where $\sigma^2_{n}$ is the thermal noise power, and $g_{b_k,k}(\theta_{b_k})$ is the large-scale fading coefficient between BS $b_k$ and user $k$. The large-scale fading coefficient includes path loss, shadow fading, and antenna gain, with the antenna gain being a function of the antenna tilt $\theta_{b_k}$ of BS $b_k$. The BO objective function $f(\mathbf{x})$ can then be expressed as the average of the average capacity
\begin{align}
    f(\mathbf{x})\! =\! \mathbbm{E}_{h}\left[\frac{\lambda_{\text{GU}}}{\lvert \mathcal{U} \rvert}\!\cdot\! \sum_{k \in \mathcal{U}} R_k (h)\! +\! \frac{1\!-\!\lambda_{\text{GU}}}{\lvert \mathcal{G} \rvert}\! \cdot\! \sum_{k \in \mathcal{G}} R_k (h) \right],
    \label{eq:CUAV_objective_func}
\end{align}
which is averaged over the fading channel $h$ as $\eqref{eq:CUAV_objective_func}$, the parameter $\lambda_{\text{GU}}$, with $0 \leq \lambda_{\text{GU}} \leq 1$ balances the capacity of the GUs and of the UAVs. In the following experiments, $\lambda_{\text{GU}}$ is set to $0.7$.

The parameters $\mathbf{x}=[\mathbf{p},\mathbf{\Theta}]$, collecting all the power $\{p_b\}_{b \in \mathcal{B}}$ and all the antenna tilts $\{\theta_b\}_{b\in \mathcal{B}}$ are optimized using BO at a centralized unit. Specifically, as in \cite{benzaghta2023designing}, at BO iteration $t$, we update the parameters $(p_b, \theta_b)$ for a base station $b$ while maintaining the parameters for the other base stations fixed. The maximization of the acquisition function is done via a grid search evaluated over 100 random candidate points. In this experiment, we additionally consider a random search (RS) method, whereby a single candidate point is randomly sampled from the input space at each iteration $t$. This method provides a reference to evaluate the effectiveness of an active selection of iterates. 

For each candidate solution $\mathbf{x}$ selected by BO, the observation $y$ is obtained by using  $n_{\text{ch}}$ independent channel realizations $[h_{b_k,k}^{(1)}, \ldots,  h_{b_k,k}^{(n_{\text{ch}})}]$ per link between UE $k$ and BS $b_k$ as 
\begin{align}
    \!\!y\! =\! \frac{1}{n_{\text{ch}}}\!\left(\!\!\frac{\lambda_{\text{GU}}}{\lvert \mathcal{U} \rvert}\!\cdot\! \sum_{k \in \mathcal{U}}\!\sum_{i=1}^{n_{\text{ch}}}\! R_k(h_{b_k,k}^{(i)})\! +\! \frac{1\!-\!\lambda_{\text{GU}}}{\lvert \mathcal{G} \rvert}\! \cdot\!  \sum_{k \in \mathcal{G}}\!\sum_{i=1}^{n_{\text{ch}}}\! R_k(h_{b_k,k}^{(i)})\!\! \right)\!,\!\!\!\!\!\!\!
    \label{eq:CUAV_objective_func_nch}
\end{align}
which is an empirical and thus noisy estimate of the average capacity \eqref{eq:CUAV_objective_func}.
We use 50 random uniformly distributed points $\mathbf{x}$ to initialize the GP model. The target miscoverage level $\alpha$ for all CP-based schemes is set to $0.25$.
\begin{figure*}[t]
    \centering
    \begin{subfigure}[b]{0.33\textwidth} 
        \centering
        \includegraphics[width=\textwidth]{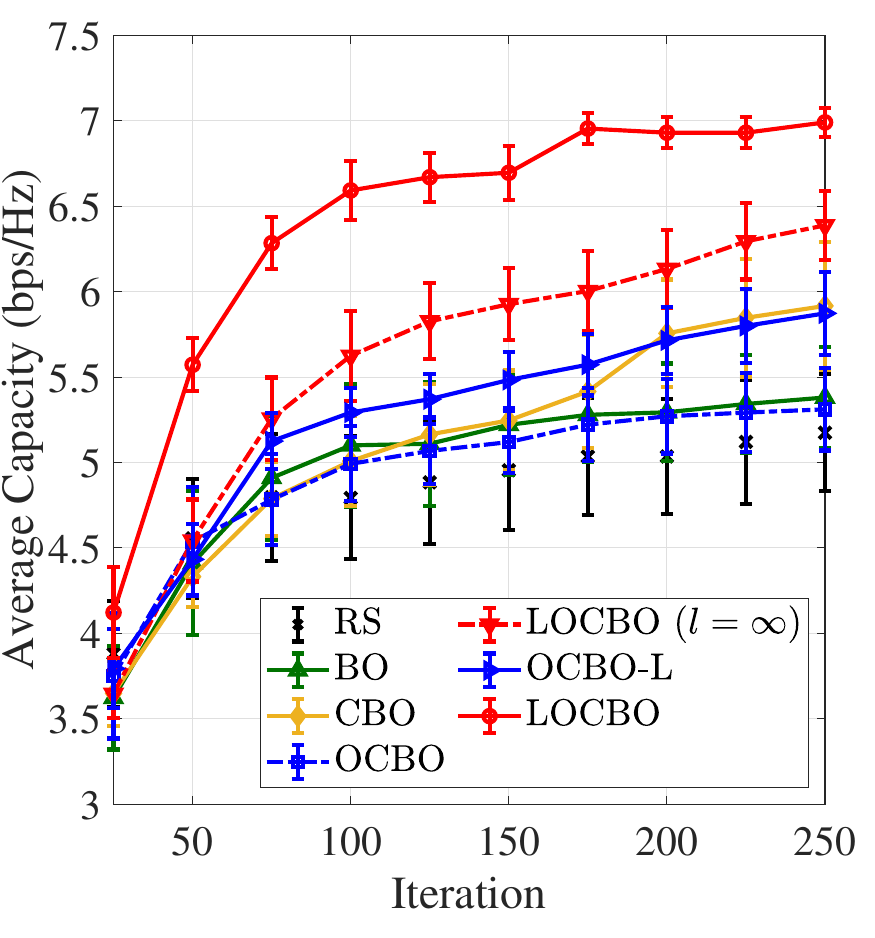}
        \caption{}
        \label{fig:CUAV_per_iter}
    \end{subfigure}
    \hfill
    \hspace{-0.02\textwidth}
    \begin{subfigure}[b]{0.33\textwidth}
        \centering
        \includegraphics[width=\textwidth]{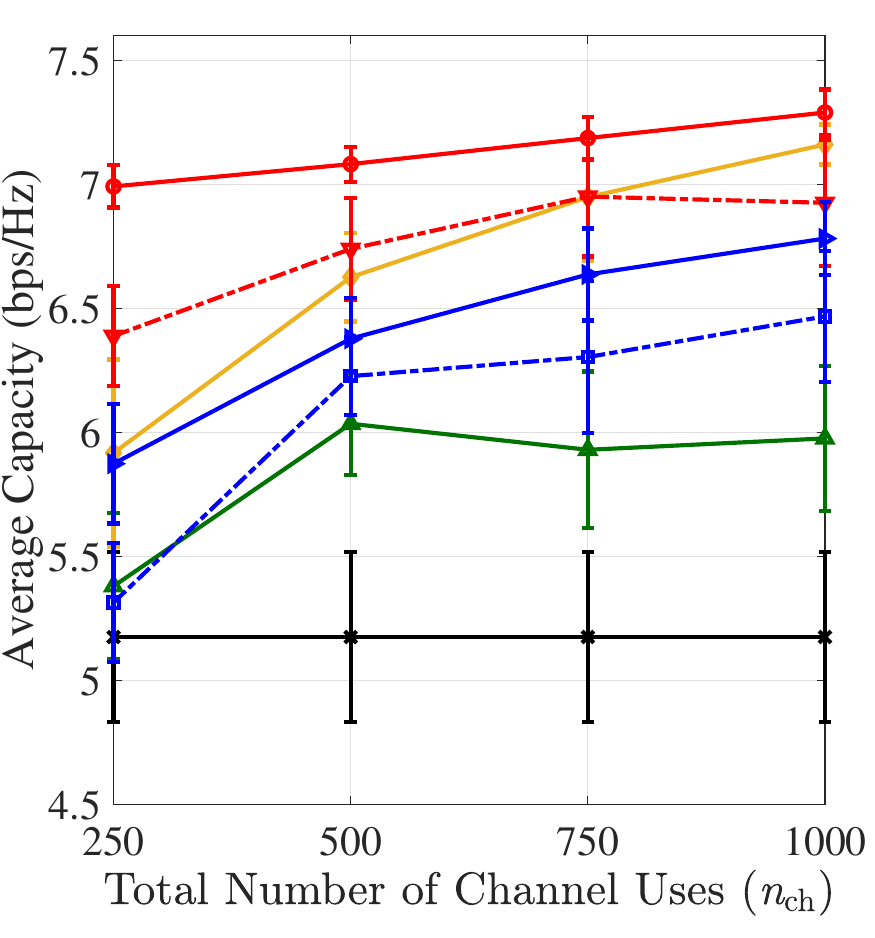}
        \caption{}
        \label{fig:CUAV_per_nch}
    \end{subfigure}
    \hfill
    \hspace{-0.02\textwidth}
    \begin{subfigure}[b]{0.33\textwidth}
        \centering
        \includegraphics[width=\textwidth]{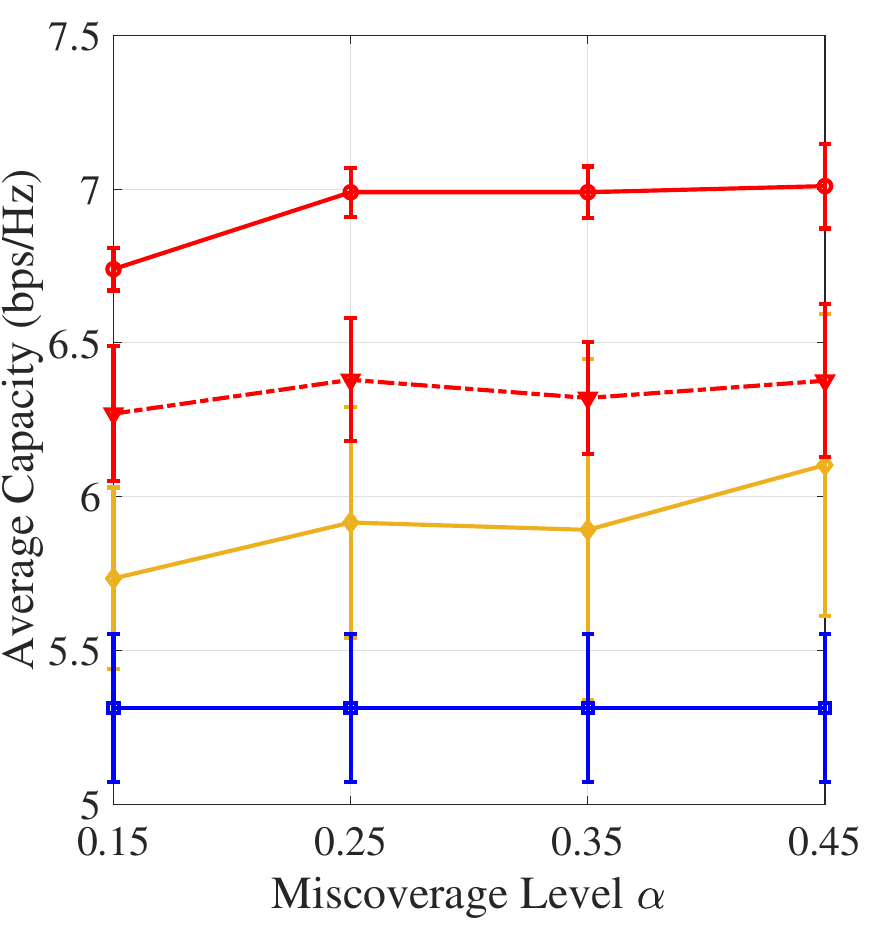}
         \caption{}
        \label{fig:CUAV_per_alpha}
    \end{subfigure}
    \caption{(a) Average capacity \eqref{eq:CUAV_objective_func} as a function of the optimization horizon $T$ for the radio resource management problem. (b) Average capacity \eqref{eq:CUAV_objective_func} as a function of the number of channel realizations $n_{\text{ch}}$ for the radio resource management problem. (c) Average capacity \eqref{eq:CUAV_objective_func} as a function of the target miscoverage level $\alpha$ for the radio resource management problem. All figures show mean and 70\% confidence interval.}
    \label{fig:overall}
    \vspace{-12pt}
\end{figure*}

In Figure \ref{fig:CUAV_per_iter}, we report the average capacity as a function of the number of optimization iterations $T$, when a single channel realization \(n_{\text{ch}} = 1\) is used to estimate the average capacity in \eqref{eq:CUAV_objective_func_nch}. All BO schemes consistently outperform RS, validating the benefits of model-based optimization via BO. Among the CP-based BO schemes, LOCBO yields the highest average capacity. Additionally, in this setting, localization provides substantial performance improvements. LOCBO with \(l =1/3\) provides roughly a 10\% gain in average capacity compared to LOCBO with \(l = \infty\), as well as a 20\% gain compared to CBO and OCBO. 

To investigate the impact of the observation noise, we now study the performance as a function of the number of channel realizations \( n_{\text{ch}} \) used to estimate the  average capacity. A larger $n_{\text{ch}}$ yields a smaller statistical error in the estimate \eqref{eq:CUAV_objective_func_nch}. In Figure \ref{fig:CUAV_per_nch}, we report the average capacity after \( T = 250 \) BO rounds when \( n_{\text{ch}} = 1, 2, 3, \text{ and } 4 \) channel realizations are used per round. Given that, a larger number of samples \( n_{\text{ch}} \) provides a better estimate of the unknown objective, the average capacity achieved by the BO schemes increases as $n_{\text{ch}}$ grows. CP-based schemes consistently demonstrate superior performance compared to standard BO, with LOCBO, using length scale \( l = 1/3\), outperforming all other baselines.

Finally, in Figure \ref{fig:CUAV_per_alpha}, we study the effect of miscoverage level $\alpha$ on the average capacity. Across all values of $\alpha$, LOCBO consistently outperforms the baseline methods in terms of maximizing average capacity. Generally, at higher values of $\alpha$, all schemes demonstrate improved performance compared to $\alpha = 0.15$, although all schemes are considered to be robust to the choice of hyperparameter $\alpha$.

\section{Conclusion}
\label{sec:Conclusion}
In this work, we propose LOCBO, a novel BO algorithm that leverages online CP calibration to mitigate the effect of GP misspecification on BO's performance. Unlike previous approaches based on CP,  LOCBO applies online CP to calibrate the GP likelihood function, while accounting for observation noise by incorporating a denoising step to produce a calibrated  GP posterior.  LOCBO builds on recent advancements in localized online CP to perform input-dependent calibration of the GP likelihood and provide a more uniform level of coverage across the input space. In this way, LOCBO is shown to provide strong performance guarantees under minimal  assumptions about the symmetry of the distribution of the observation noise. Empirical evaluations on both synthetic objective functions and real-world engineering problems demonstrate the advantages of localized calibration and the superior performance of LOCBO compared to existing CP-based BO methods. 

For future work, several promising directions could enhance and broaden the applicability of LOCBO. One potential direction is developing an online approach for updating LOCBO’s hyperparameters that automatically controls the degree of localization \cite{angelopoulos2024conformal}. Additionally, extensions to multi-objective and multi-fidelity optimization problems \cite{zhang2024multi}, as well as improvements in scalability for higher-dimensional search spaces \cite{eriksson2019scalable}, would be of particular interest.
\bibliographystyle{IEEEtran}
\bibliography{LOCBO}

\begin{thebibliography}{10}
\providecommand{\url}[1]{#1}
\csname url@samestyle\endcsname
\providecommand{\newblock}{\relax}
\providecommand{\bibinfo}[2]{#2}
\providecommand{\BIBentrySTDinterwordspacing}{\spaceskip=0pt\relax}
\providecommand{\BIBentryALTinterwordstretchfactor}{4}
\providecommand{\BIBentryALTinterwordspacing}{\spaceskip=\fontdimen2\font plus
\BIBentryALTinterwordstretchfactor\fontdimen3\font minus \fontdimen4\font\relax}
\providecommand{\BIBforeignlanguage}[2]{{%
\expandafter\ifx\csname l@#1\endcsname\relax
\typeout{** WARNING: IEEEtran.bst: No hyphenation pattern has been}%
\typeout{** loaded for the language `#1'. Using the pattern for}%
\typeout{** the default language instead.}%
\else
\language=\csname l@#1\endcsname
\fi
#2}}
\providecommand{\BIBdecl}{\relax}
\BIBdecl

\bibitem{wang2022tight}
Z.~Wang, V.~Y. Tan, and J.~Scarlett, ``Tight regret bounds for noisy optimization of a {Brownian} motion,'' \emph{IEEE Transactions on Signal Processing}, vol.~70, pp. 1072--1087, 2022.

\bibitem{zhang2023bayesian}
Y.~Zhang, O.~Simeone, S.~T. Jose, L.~Maggi, and A.~Valcarce, ``Bayesian and multi-armed contextual meta-optimization for efficient wireless radio resource management,'' \emph{IEEE Transactions on Cognitive Communications and Networking}, vol.~9, no.~5, pp. 1282--1295, 2023.

\bibitem{snoek2014input}
J.~Snoek, K.~Swersky, R.~Zemel, and R.~Adams, ``Input warping for {Bayesian} optimization of non-stationary functions,'' in \emph{International conference on machine learning}, Beijing, China, 2014.

\bibitem{bodin2020modulating}
E.~Bodin, M.~Kaiser, I.~Kazlauskaite, Z.~Dai, N.~Campbell, and C.~H. Ek, ``Modulating surrogates for {B}ayesian optimization,'' in \emph{International Conference on Machine Learning}, 2020.

\bibitem{bogunovic2021misspecified}
I.~Bogunovic and A.~Krause, ``Misspecified {Gaussian} process bandit optimization,'' \emph{Advances in neural information processing systems}, 2021.

\bibitem{griffiths2021achieving}
R.-R. Griffiths, A.~A. Aldrick, M.~Garcia-Ortegon, V.~Lalchand \emph{et~al.}, ``Achieving robustness to aleatoric uncertainty with heteroscedastic {Bayesian} optimisation,'' \emph{Machine Learning: Science and Technology}, vol.~3, no.~1, p. 015004, 2021.

\bibitem{9212578}
R.~Guzman, R.~Oliveira, and F.~Ramos, ``Heteroscedastic {Bayesian} optimisation for stochastic model predictive control,'' \emph{IEEE Robotics and Automation Letters}, vol.~6, no.~1, pp. 56--63, 2021.

\bibitem{schulz2016quantifying}
E.~Schulz, M.~Speekenbrink, J.~M. Hern{\'a}ndez-Lobato, Z.~Ghahramani, and S.~J. Gershman, ``Quantifying mismatch in {Bayesian} optimization,'' in \emph{Nips workshop on bayesian optimization: Black-box optimization and beyond}, 2016.

\bibitem{shahriari2015taking}
B.~Shahriari, K.~Swersky, Z.~Wang, R.~P. Adams, and N.~De~Freitas, ``Taking the human out of the loop: {A} review of {Bayesian} optimization,'' \emph{Proceedings of the IEEE}, vol. 104, no.~1, pp. 148--175, 2015.

\bibitem{neiswanger2021uncertainty}
W.~Neiswanger and A.~Ramdas, ``Uncertainty quantification using martingales for misspecified {Gaussian} processes,'' in \emph{Algorithmic learning theory}, 2021.

\bibitem{stanton2023bayesian}
S.~Stanton, W.~Maddox, and A.~G. Wilson, ``Bayesian optimization with conformal prediction sets,'' in \emph{International Conference on Artificial Intelligence and Statistics}, Valencia, Spain, 2023.

\bibitem{deshpande2024online}
S.~Deshpande, C.~Marx, and V.~Kuleshov, ``Online calibrated and conformal prediction improves {B}ayesian optimization,'' in \emph{International Conference on Artificial Intelligence and Statistics}, Valencia, Spain, 2024.

\bibitem{zecchin2024localized}
M.~Zecchin and O.~Simeone, ``Localized adaptive risk control,'' in \emph{Advances in neural information processing systems}, Vancouver, Canada, 2024.

\bibitem{albahar2022physics}
A.~AlBahar, I.~Kim, X.~Wang, and X.~Yue, ``Physics-constrained bayesian optimization for optimal actuators placement in composite structures assembly,'' \emph{IEEE Transactions on Automation Science and Engineering}, vol.~20, no.~4, pp. 2772--2783, 2022.

\bibitem{marmin2018warped}
S.~Marmin, D.~Ginsbourger, J.~Baccou, and J.~Liandrat, ``Warped gaussian processes and derivative-based sequential designs for functions with heterogeneous variations,'' \emph{SIAM/ASA Journal on Uncertainty Quantification}, vol.~6, no.~3, pp. 991--1018, 2018.

\bibitem{nikoloska2022bayesian}
I.~Nikoloska and O.~Simeone, ``Bayesian active meta-learning for black-box optimization,'' in \emph{2022 IEEE 23rd International Workshop on Signal Processing Advances in Wireless Communication (SPAWC)}, Oulu, Finland, 2022.

\bibitem{volpp2019meta}
M.~Volpp, L.~P. Fr{\"o}hlich, K.~Fischer, A.~Doerr, S.~Falkner, F.~Hutter, and C.~Daniel, ``Meta-learning acquisition functions for transfer learning in {Bayesian} optimization,'' \emph{arXiv preprint arXiv:1904.02642}, 2019.

\bibitem{rothfuss2023meta}
J.~Rothfuss, C.~Koenig, A.~Rupenyan, and A.~Krause, ``Meta-learning priors for safe {B}ayesian optimization,'' in \emph{Conference on Robot Learning}, Atlanta, USA, 2023.

\bibitem{sugiyama2005active}
M.~Sugiyama, ``Active learning for misspecified models,'' \emph{Advances in neural information processing systems}, vol.~18, 2005.

\bibitem{fudenberg2017active}
D.~Fudenberg, G.~Romanyuk, and P.~Strack, ``Active learning with a misspecified prior,'' \emph{Theoretical Economics}, vol.~12, no.~3, pp. 1155--1189, 2017.

\bibitem{svendsen2020active}
D.~H. Svendsen, L.~Martino, and G.~Camps-Valls, ``Active emulation of computer codes with gaussian processes--application to remote sensing,'' \emph{Pattern Recognition}, vol. 100, p. 107103, 2020.

\bibitem{fernandez2020adaptive}
F.~L. Fern{\'a}ndez, L.~Martino, V.~Elvira, D.~Delgado, and J.~L{\'o}pez-Santiago, ``Adaptive quadrature schemes for bayesian inference via active learning,'' \emph{IEEE Access}, vol.~8, pp. 208\,462--208\,483, 2020.

\bibitem{vovk2005algorithmic}
V.~Vovk, A.~Gammerman, and G.~Shafer, \emph{Algorithmic learning in a random world}.\hskip 1em plus 0.5em minus 0.4em\relax Springer, 2005, vol.~29.

\bibitem{angelopoulos2021gentle}
A.~N. Angelopoulos and S.~Bates, ``A gentle introduction to conformal prediction and distribution-free uncertainty quantification,'' \emph{arXiv preprint arXiv:2107.07511}, 2021.

\bibitem{gibbs2021adaptive}
I.~Gibbs and E.~Candes, ``Adaptive conformal inference under distribution shift,'' \emph{Advances in Neural Information Processing Systems}, 2021.

\bibitem{angelopoulos2024online}
A.~N. Angelopoulos, R.~F. Barber, and S.~Bates, ``Online conformal prediction with decaying step sizes,'' \emph{arXiv preprint arXiv:2402.01139}, 2024.

\bibitem{feldman2022achieving}
S.~Feldman, L.~Ringel, S.~Bates, and Y.~Romano, ``Achieving risk control in online learning settings,'' \emph{arXiv preprint arXiv:2205.09095}, 2022.

\bibitem{tibshirani2019conformal}
R.~J. Tibshirani, R.~Foygel~Barber, E.~Candes, and A.~Ramdas, ``Conformal prediction under covariate shift,'' in \emph{Advances in neural information processing systems}, Vancouver, Canada, 2019.

\bibitem{guan2019conformal}
L.~Guan, ``Conformal prediction with localization,'' \emph{arXiv preprint arXiv:1908.08558}, 2019.

\bibitem{gibbs2022conformal}
I.~Gibbs and E.~Cand{\`e}s, ``Conformal inference for online prediction with arbitrary distribution shifts,'' \emph{arXiv preprint arXiv:2208.08401}, 2022.

\bibitem{hore2023conformal}
R.~Hore and R.~F. Barber, ``Conformal prediction with local weights: randomization enables local guarantees,'' \emph{arXiv preprint arXiv:2310.07850}, 2023.

\bibitem{lee2017hierarchically}
B.-J. Lee, J.~Lee, and K.-E. Kim, ``Hierarchically-partitioned gaussian process approximation,'' in \emph{Artificial Intelligence and Statistics}.\hskip 1em plus 0.5em minus 0.4em\relax PMLR, 2017, pp. 822--831.

\bibitem{eriksson2019scalable}
D.~Eriksson, M.~Pearce, J.~Gardner, R.~D. Turner, and M.~Poloczek, ``Scalable global optimization via local {B}ayesian optimization,'' in \emph{Advances in neural information processing systems}, Vancouver, Canada, 2019.

\bibitem{surjanovic2013virtual}
S.~Surjanovic and D.~Bingham, ``Virtual library of simulation experiments: {Test} functions and datasets,'' in \emph{Simon Fraser University, Burnaby, BC, Canada, accessed May}, vol.~13, 2013.

\bibitem{benzaghta2023designing}
M.~Benzaghta, G.~Geraci, D.~L{\'o}pez-P{\'e}rez, and A.~Valcarce, ``Designing cellular networks for uav corridors via bayesian optimization,'' in \emph{IEEE Global Communications Conference}, Kuala Lumpur, Malaysia, 2023, pp. 4552--4557.

\bibitem{williams2006gaussian}
C.~K. Williams and C.~E. Rasmussen, \emph{Gaussian processes for machine learning}.\hskip 1em plus 0.5em minus 0.4em\relax MIT press Cambridge, MA, 2006, vol.~2, no.~3.

\bibitem{srinivas2009gaussian}
N.~Srinivas, A.~Krause, S.~M. Kakade, and M.~Seeger, ``Gaussian process optimization in the bandit setting: {N}o regret and experimental design,'' \emph{arXiv preprint arXiv:0912.3995}, 2009.

\bibitem{wilson2017reparameterization}
J.~T. Wilson, R.~Moriconi, F.~Hutter, and M.~P. Deisenroth, ``The reparameterization trick for acquisition functions,'' \emph{arXiv preprint arXiv:1712.00424}, 2017.

\bibitem{kivinen2004online}
J.~Kivinen, A.~J. Smola, and R.~C. Williamson, ``Online learning with kernels,'' \emph{IEEE transactions on signal processing}, vol.~52, no.~8, pp. 2165--2176, 2004.

\bibitem{guille2024conformal}
C.~Guille-Escuret and E.~Ndiaye, ``From conformal predictions to confidence regions,'' \emph{arXiv preprint arXiv:2405.18601}, 2024.

\bibitem{snoek2012practical}
J.~Snoek, H.~Larochelle, and R.~P. Adams, ``Practical {Bayesian} optimization of machine learning algorithms,'' in \emph{Advances in neural information processing systems}, Lake Tahoe, USA, 2012.

\bibitem{frazier2018tutorial}
P.~I. Frazier, ``A tutorial on {B}ayesian optimization,'' \emph{arXiv preprint arXiv:1807.02811}, 2018.

\bibitem{welling2011bayesian}
M.~Welling and Y.~W. Teh, ``Bayesian learning via stochastic gradient langevin dynamics,'' in \emph{International Conference on Machine Learning}, Bellevue, USA, 2011.

\bibitem{wang2017max}
Z.~Wang and S.~Jegelka, ``Max-value entropy search for efficient {Bayesian} optimization,'' in \emph{International Conference on Machine Learning}, Sydney, Australia, 2017.

\bibitem{specification3gpp}
3GPP, ``{Study on channel model for frequencies from 0.5 to 100 GHz (Release 16)},'' {3rd Generation Partnership Project (3GPP)}, TR 38.901, Dec. 2019.

\bibitem{angelopoulos2024conformal}
A.~Angelopoulos, E.~Candes, and R.~J. Tibshirani, ``Conformal pid control for time series prediction,'' in \emph{Advances in neural information processing systems}, Vancouver, Candada, 2024.

\bibitem{zhang2024multi}
Y.~Zhang, S.~Park, and O.~Simeone, ``Multi-fidelity {Bayesian} optimization with across-task transferable max-value entropy search,'' \emph{arXiv preprint arXiv:2403.09570}, 2024.

\end{thebibliography}
\appendix
\section{Appendix}
\subsection{Closed-Form Expression of LOCBO's Posterior}
\label{appendix:LOCBO_decomp}
The following lemma describes the result of the integration \eqref{eq:LOCBO_decomp}.
\begin{lemma}[LOCBO's posterior distribution $p_{\alpha}^{\text{\tiny{LOCBO}}}
\label{lemma: LOCBO's calibrated posterior}
(f(\mathbf{x})|\mathcal{D}_t)$] The posterior distribution of LOCBO is expressed in \eqref{eq: LOCBO rectified posterior}.
\begin{figure}[h!]
\vspace{-0.2cm}
\begin{align}
    \label{eq: LOCBO rectified posterior}
    &p^{\textnormal{\tiny{LOCBO}}}_\alpha(f(\mathbf{x})|\mathcal{D}_t)\!=\! 
    \frac{1-\alpha}{2a(U_{t+1}(\mathbf{x}|\mathcal{D}_t)\!\!-\!\!L_{t+1}(\mathbf{x}|\mathcal{D}_t))}
    \nonumber\\&\!\cdot\! \left[ \erf\left( \frac{-f(\mathbf{x}) + a U_{t+1}(\mathbf{x}|\mathcal{D}_t) + b}{\sqrt{2}\tilde{\sigma}(\mathbf{x}|\mathcal{D}_{t+1}^{(\mathbf{x},y')})} \right) \right. 
    -\nonumber\\& \left. \erf\left( \frac{-f(\mathbf{x}) + a L_{t+1}(\mathbf{x}|\mathcal{D}_t) + b}{\sqrt{2}\tilde{\sigma}(\mathbf{x}|\mathcal{D}_{t+1}^{(\mathbf{x},y')})} \right) \right] \nonumber\\
    &\nonumber + \frac{\alpha}{4\pi \lambda_t(\mathbf{x}) \tilde{\sigma}(\mathbf{x}|\mathcal{D}_{t+1}^{(\mathbf{x},y')}) \tilde{\sigma}(\mathbf{x}|\mathcal{D}_t)} 
    \sqrt{\frac{\pi}{A^2 + C^2}}\\ &\cdot \exp\left( -\frac{A^2 C^2}{(A^2 + C^2)(B - D)^2} \right) \nonumber\\
    &\nonumber \cdot \left[ 
2 - \erf\left( \frac{-A^2 B - C^2 D + (A^2 + C^2) U_{t+1}(\mathbf{x}|\mathcal{D}_t)}{\sqrt{A^2 + C^2}} \right) 
\right. \\
& \left. - \erf\left( \frac{-A^2 B - C^2 D + (A^2 + C^2) L_{t+1}(\mathbf{x}|\mathcal{D}_t)}{\sqrt{A^2 + C^2}} \right) 
\right],
\end{align}
\end{figure}

\noindent {where} $\mathcal{D}_{t+1}^{(\mathbf{x},y')} = \mathcal{D}_t \cup \{(\mathbf{x},y')\}$, 
$\tilde{\sigma}^2(\mathbf{x}|\mathcal{D}_{t+1}^{(\mathbf{x},y')}) = \sigma^2(\mathbf{x}|\mathcal{D}_{t+1}^{(\mathbf{x},y')}) + \sigma^2$,
$\tilde{\sigma}^2(\mathbf{x}|\mathcal{D}_t) = \sigma^2(\mathbf{x}|\mathcal{D}_t) + \sigma^2$, $A = \frac{a^2}{2\tilde{\sigma}^2(\mathbf{x}|\mathcal{D}_{t+1}^{(\mathbf{x},y')})},\, B = \frac{f(\mathbf{x})-b}{a},\, C = \frac{1}{2\tilde{\sigma}^2(\mathbf{x}|\mathcal{D}_t)},\, D = \mu(\mathbf{x}|\mathcal{D}_t)$, denoting as $a$ the last element of the $1 \times (t+1)$ vector $\mu' = \mathbf{k}'(\mathbf{x})^\top(\mathbf{K}'(\mathbf{X})+\sigma^2\mathbf{I}_{t+1})^{-1}$ given $\mathbf{k}'(\mathbf{x}) = [\mathbf{k}(\mathbf{x});k(\mathbf{x},\mathbf{x})]^\top$ and $\mathbf{K}'(\mathbf{X}) = [\mathbf{K}(\mathbf{X}), \mathbf{k}'(\mathbf{x})\nonumber;\mathbf{k}'(\mathbf{x})^\top, \linebreak[0] k(\mathbf{x},\mathbf{x})]$; while defining as $b = \mu'_{1:t}\mathbf{y}$ with  $\mu'_{1:t}$ being the $1 \times t$ vector composed of the first $t$ elements of $\mu'$, i.e., $\mu'_{1:t} = [\mu'_1, ..., \mu'_t]$ with $\mu_i'$ being the $i$-th element of $\mu'$. 
\end{lemma}

\section{Proofs}
\subsection{Proof of Lemma \ref{lemma:long-run_noisy}}
The proof follows from \cite[Theorem 2]{zecchin2024localized} by noting that, for all $t\geq 1$ and any pair $(y,\mathbf{x})$, LOCBO's NC score in \eqref{eq:locbo_nc_score} satisfies 
\begin{align}
    \label{eq:lower_bound_single_step}
    s_t(\mathbf{x},y)=2 Q\left(\frac{|y-\mu(\mathbf{x}|\mathcal{D}_{t-1})|}{\tilde{\sigma}(\mathbf{x}|\mathcal{D}_{t-1})}\right)\leq 2,
\end{align}
and the miscoverage loss $ \mathbbm{1}\{y \notin  \Gamma_{t}^{\text{\tiny{LOCBO}}}(\mathbf{x})\}\leq 1$.

\subsection{Proof of Lemma \ref{lemma:long-run_obj}}
Given observation $y$ as in \eqref{eq:noise_modelling} with noise $\xi(\mathbf{x})$ satisfying Assumption  \eqref{ass:balanced_noise}, given any interval predictor $\Gamma(\mathbf{x})=[L(\mathbf{x}), U(\mathbf{x})]$ it holds
\begin{align}
    \Pr[f(\mathbf{x})\notin \Gamma(\mathbf{x})]\leq \frac{\Pr[y\notin \Gamma(\mathbf{x})]}{b_\xi}
    \label{eq:f_x_coverage_y_coverage}
\end{align}
where the expectation is over the noise $\xi(\mathbf{x})$ \cite[Proposition 2.3]{guille2024conformal}.

For any query sequence $\{\mathbf{x}_t\}^T_{t=1}$ and independent observations  $\{y_t\}^T_{t=1}$ with $y_t$ as in \eqref{eq:noise_modelling}, Lemma \ref{lemma:long-run_noisy} guarantees that for any hyperparameter $\lambda>0$ and any learning rate sequence $\eta_t=\eta_1 t^{-1/2}< 1/\lambda$ with $\eta_1>0$, LOCBO's prediction sets $\{\Gamma_{t}^{\textnormal{\tiny{LOCBO}}}(\mathbf{x}_t|\mathcal{D}_{t-1})\}_{t\geq 1}$ satisfy the deterministic coverage condition
       \begin{align}
			\frac{1}{T}\sum^T_{t=1} \mathbbm{1}\{y_t\notin \Gamma_{t}^{\textnormal{\tiny{LOCBO}}}(\mathbf{x}_t|\mathcal{D}_{t-1}) \}\leq  \alpha+ \frac{\beta}{\sqrt{T}} +{\color{black}\kappa},
		\end{align}
        with $\beta=\frac{2}{\eta_1 }+\frac{4 \sqrt{\rho{\color{black}\kappa D}}}{\eta_1\lambda}+2(2\kappa+1)$.
Taking the expectation with respect to the noise sequence  $\{\xi(\mathbf{x}_t)\}_{t\geq 1}$ yields
   \begin{align}
			\frac{1}{T}\sum^T_{t=1} \Pr\left[y_t\notin \Gamma_{t}^{\textnormal{\tiny{LOCBO}}}(\mathbf{x}_t|\mathcal{D}_{t-1}) \right]\leq  \alpha+ \frac{\beta}{\sqrt{T}} +{\color{black}\kappa}.
		\end{align}
Inequality \eqref{eq:f_x_coverage_y_coverage}, applied to the summation terms, yields the final result 
  \begin{align}
			\frac{1}{T}\sum^T_{t=1} \Pr\left[f(\mathbf{x}_t)\notin \Gamma_{t}^{\textnormal{\tiny{LOCBO}}}(\mathbf{x}_t|\mathcal{D}_{t-1}) \right]\leq  \frac{1}{b_\xi}\left(\alpha+ \frac{\beta}{\sqrt{T}} +{\color{black}\kappa}\right).
		\end{align}
\subsection{Proof of Lemma \ref{lemma: LOCBO's calibrated posterior}}
Let $\mathbf{X}' = [\mathbf{x}_1,\ldots \mathbf{x}_t,\mathbf{x}]^\top \in \mathbbm{R}^{(t+1)\times d}$ , $\mathbf{y} = [y_1,\ldots, y_t]^\top \in \mathbbm{R}^{t}$, and $\mathcal{D}_{t+1}^{(\mathbf{x},y')} = \mathcal{D}_t\cup \{(\mathbf{x}, y')\}$. We can write joint distribution of the observation vector $\mathbf{y}$, the observation value $y'$ at the $\mathbf{x}$, and the objective function $f(\mathbf{x})$ at $\mathbf{x}$ under the GP prior as 
\begin{align}
\begin{bmatrix}
\mathbf{y}  \\
y'  \\
f(\mathbf{x})
\end{bmatrix} \sim \mathcal{N}\left(0,
\begin{bmatrix}
\mathbf{K}(\mathbf{X}', \mathbf{X}') + \sigma^2\mathbf{I}_{t+1} & \mathbf{k}(\mathbf{X}', \mathbf{x}) \\
\mathbf{k}(\mathbf{x}, \mathbf{X}') & \mathbf{k}(\mathbf{x}, \mathbf{x})
\end{bmatrix}\right).
\end{align}
Therefore, the random variable $f(\mathbf{x})$ conditioned on $\mathcal{D}_{t+1}^{(\mathbf{x},y')}$ follows a Gaussian distribution with variance $\tilde{\sigma}^2(\mathbf{x}|\mathcal{D}_{t+1}^{(\mathbf{x},y')}) = \mathbf{k}(\mathbf{x}, \mathbf{x}) - \mathbf{k}(\mathbf{x}, \mathbf{X}')(\mathbf{k}(\mathbf{X}', \mathbf{X}')+\sigma^2\mathbf{I}_{t+1})^{-1}\mathbf{k}(\mathbf{X}', \mathbf{x})$ and mean $ay'+b$. The likelihood $p(y'|\mathbf{x},\mathcal{D}_t)$ also follows a Gaussian distribution with variance $\tilde{\sigma}^2(\mathbf{x}|\mathcal{D}_t)$ and mean $\mu(\mathbf{x}|\mathcal{D}_t)$. Once we calculate the integration in \eqref{eq:LOCBO_decomp}, LOCBO's posterior is derived as in \eqref{eq: LOCBO rectified posterior}.
\subsection{Proof of Theorem \ref{th:utility_guarantee}}
Pick  $x\in\mathcal{X}$ and $f'\in \mathbbm{R}$ such that the c.d.f of the calibrated likelihood satisfies
\begin{align} 
\label{eq:online conformalized app}	\int^{f'}_{-\infty}p^{\text{\tiny{LOCBO}}}_\alpha(y|\mathbf{x},\mathcal{D}_t)dy\geq\frac{\alpha}{2}.
\end{align}
The average utility at $\mathbf{x}$ estimated via LOCBO's posterior can be upper bounded as follows
\allowdisplaybreaks
\begin{align}
&a^{\text{\tiny{LOCBO}}}(\mathbf{x}|\mathcal{D}_t) = \int  u(\mathbf{x}, f(\mathbf{x}), \mathcal{D}_t)p^{\text{\tiny{LOCBO}}}_\alpha(f(\mathbf{x})|\mathcal{D}_t)df(\mathbf{x}) \\
&\!\!\!\leq u(\mathbf{x}, f', \mathcal{D}_t)\int_{-\infty}^{f'}p^{\text{\tiny{LOCBO}}}_\alpha(f(\mathbf{x})|\mathcal{D}_t)df(\mathbf{x})\\
&\!\!\!=\! u(\mathbf{x}, f', \mathcal{D}_t)\!\!\int_{-\infty}^{f'}\!\!\int^{\infty}_{-\infty} \!\!\!\!p(f(\mathbf{x})|\mathcal{D}_{t+1}^{(\mathbf{x},y')})p^{\text{\tiny{LOCBO}}}_\alpha(y|\mathbf{x},\mathcal{D}_t)dydf(\mathbf{x})\\
	&\!\!\!=\! u(\mathbf{x}, f', \mathcal{D}_t)\!\!\int^{\infty}_{-\infty}\!\!\!\!p^{\text{\tiny{LOCBO}}}_\alpha(y|\mathbf{x},\mathcal{D}_t)\!\!\int_{-\infty}^{f'}\!\!\!\!p(f(\mathbf{x})|\mathcal{D}_{t+1}^{(\mathbf{x},y')})df(\mathbf{x})dy\\
     &\!\!\!\leq\! u(\mathbf{x}, f', \mathcal{D}_t)\!\!\int_{-\infty}^{f'}\!\!\!\!p^{\text{\tiny{LOCBO}}}_\alpha(y|\mathbf{x},\mathcal{D}_t)\!\!\int_{-\infty}^{f'}\!\!\!\!p(f(\mathbf{x})|\mathcal{D}_{t+1}^{(\mathbf{x},y')})df(\mathbf{x})dy\\
	&\!\!\!\leq\! u(\mathbf{x}, f', \mathcal{D}_t)\!\! \int_{-\infty}^{f'}\!\!\!\!p^{\text{\tiny{LOCBO}}}_\alpha(y|\mathbf{x},\mathcal{D}_t)\epsilon dy\\
	&\!\!\!\leq\! u(\mathbf{x}, f', \mathcal{D}_t)\frac{\alpha\epsilon}{2}.
\end{align}
The first inequality follows from the bounded positive with $C=0$ and monotonicity of the utility (Assumption \ref{ass:decreasing_utility}), the second one from the fact that the c.d.f $\int_{-\infty}^{f'}p(f(\mathbf{x})|\mathcal{D}_{t+1}^{(\mathbf{x},y')})df$ is non-negative, the third inequality follows from Assumption \ref{ass:conservative_posterior}, and the last one follows from the choice of $f'$ in \eqref{eq:online conformalized app}.

The above implies that when $f'$ satisfies the condition \eqref{eq:online conformalized app} then $a^{\text{\tiny{LOCBO}}}(\mathbf{x}|\mathcal{D}_t) \leq u(\mathbf{x}, f', \mathcal{D}_t)\frac{\alpha\epsilon}{2}$.
It then follows that when $a^{\text{\tiny{LOCBO}}}(\mathbf{x}|\mathcal{D}_t)  > u(\mathbf{x}, f', \mathcal{D}_t)\frac{\alpha\epsilon}{2}$ we must have 
\begin{align} 
\int^{f'}_{-\infty}p^{\text{\tiny{LOCBO}}}_\alpha(y|\mathbf{x},\mathcal{D}_t)dy<\frac{\alpha}{2}.
\end{align}
which implies $f'\notin \Gamma_{t+1}^{\text{\tiny{LOCBO}}}(\mathbf{x}|\mathcal{D}_{t})$ by construction \eqref{eq:locbo_likelihood}.
We then conclude that for any $\mathbf{x}\in\mathcal{X}$ and $f'\in \mathbb{R}$
\begin{align}
	\Pr\left[a^{\text{\tiny{LOCBO}}}(\mathbf{x}|\mathcal{D}_t)>u(\mathbf{x}, f', \mathcal{D}_t)\frac{\alpha\epsilon}{2}\right]\leq \Pr\left[f'\notin \Gamma_{t+1}^{\text{\tiny{LOCBO}}}(\mathbf{x}|\mathcal{D}_{t})\right].
\end{align}
Fixing $\mathbf{x}=\mathbf{x}_{t+1}$ and $f'=f(\mathbf{x}_{t+1})$ we obtain
\begin{align}
	&\Pr\left[a^{\text{\tiny{LOCBO}}}(\mathbf{x}_{t+1}|\mathcal{D}_t)> u(\mathbf{x}_{t+1}, f(\mathbf{x}_{t+1}), \mathcal{D}_t)\frac{\alpha\epsilon}{2}\right]\nonumber \\  &\leq \Pr\left[f(\mathbf{x}_{t+1})\notin \Gamma_{t+1}^{\text{\tiny{LOCBO}}}(\mathbf{x}|\mathcal{D}_t)\right].
\end{align}
Summing over $t=1,\dots,T$ we obtain
\begin{align}
	&\frac{1}{T}\sum^T_{t=1}\Pr\left[(a^{\text{\tiny{LOCBO}}}(\mathbf{x}_{t}|\mathcal{D}_{t-1})> (u(\mathbf{x}_{t}, f(\mathbf{x}_{t}), \mathcal{D}_{t-1})\frac{\alpha\epsilon}{2}\right]\nonumber \\& \leq \frac{1}{T}\sum^T_{t=1}\Pr\left[f(\mathbf{x}_t)\notin \Gamma_{t}^{\text{\tiny{LOCBO}}}(\mathbf{x}_t|\mathcal{D}_{t-1})\right],
\end{align}
which together with Lemma \ref{lemma:long-run_obj} it yields the final result
\begin{align}
	&\frac{1}{T}\sum^T_{t=1}\Pr\left[a^{\text{\tiny{LOCBO}}}(\mathbf{x}_{t}|\mathcal{D}_{t-1}) > u(\mathbf{x}_{t}, f(\mathbf{x}_{t}), \mathcal{D}_{t-1})\frac{\alpha \epsilon}{2}\right]\nonumber \\&\leq  \frac{1}{b_{\xi}}\left(\alpha+ \frac{\beta}{\sqrt{T}} +{\color{black}\kappa B}\right).
\end{align}
\vfill
\end{document}